\pdfoutput=1
\documentclass[11pt]{article}

\usepackage{acl}

\usepackage{times}
\usepackage{latexsym}

\usepackage[normalem]{ulem}
\usepackage{mathrsfs}
\usepackage{amsthm}
\usepackage{amsmath}
\usepackage{amssymb, bm}
\usepackage{multirow}
\usepackage{booktabs}
\usepackage{enumitem}
\usepackage{times}
\usepackage{multirow}
\usepackage{graphicx}
\usepackage{amsfonts}
\usepackage{subcaption}
\usepackage[table]{xcolor}

\usepackage[T1]{fontenc}
\usepackage[utf8]{inputenc}

\newcommand{\tabincell}[2]{\begin{tabular}{@{}#1@{}}#2\end{tabular}}

\usepackage{microtype}

\newcommand{\secref}[1]{\S\ref{#1}}

\title{A Model-Agnostic Data Manipulation Method for \\
Persona-based Dialogue Generation}

\author{Yu Cao\textsuperscript{1}\thanks{\;\;Work was done when Yu Cao was an intern at Tencent AI Lab.}\;, Wei Bi\textsuperscript{2}\thanks{\;\;Corresponding author}\;, Meng Fang\textsuperscript{3}, Shuming Shi\textsuperscript{2}, Dacheng Tao\textsuperscript{1,4} \\
\textsuperscript{1}{School of Computer Science, The University of Sydney, Australia}\\
\textsuperscript{2}{Tencent AI Lab, Shenzhen, China}\\
\textsuperscript{3}{Eindhoven University of Technology (TU/e), Eindhoven, The Netherlands}\\
\textsuperscript{4}{JD Explore Academy, Beijing, China} \\
{\tt ycao8647@sydney.edu.au, victoriabi@tencent.com,} \\ {\tt m.fang@tue.nl, shumingshi@tencent.com, dacheng.tao@gmail.com}\\
}

\begin{document}
\maketitle
\begin{abstract}
%The ability to generate responses with consistent personas is important towards building intelligent dialogue agents. 
Towards building intelligent dialogue agents, there has been a growing interest in introducing explicit personas in generation models.
%in that generated responses need to be relevant to the dialogue context as well as maintaining consistent personas.
%Previous work has shown promising results on this task.
%in which a persona-consistent response needs to be generated conditioning both persona texts and dialogue utterances, being more complex than conventional dialogues.
%Multiple persona texts and utterances exist in one sample and some of them can be distractors for generating. Thus even strong models have difficulty posing attention to suitable personas so generating persona-irrelevant responses. 
%Besides, 
However, with limited persona-based dialogue data at hand, it may be difficult to train a dialogue generation model well.
%Our work analyzes this task from its training data directly.
We point out that the data challenges of this generation task lie in two aspects:
first, it is expensive to scale up current persona-based dialogue datasets; second, each data sample in this task is more complex to learn with than conventional dialogue data.
To alleviate the above data issues, we propose a data manipulation method, which is model-agnostic to be packed with any persona-based dialogue generation model to improve its performance.
%We show how such challenges can be addressed by data distillation and diversification.
%in data-level distillation and diversification (\textbf{D$^3$}).
The original training samples will first be distilled and thus expected to be fitted more easily.
Next, we show various effective ways that can diversify such easier distilled data.
%Our method is model-agnostic and packed with 
A given base model will then be trained via the constructed data curricula, i.e. first on augmented distilled samples and then on original ones.
Experiments illustrate the superiority of our method with two strong base dialogue models (Transformer encoder-decoder and GPT2).
%on various automatic metrics and human evaluation. 
\end{abstract}

\section{Introduction}
\begin{figure}[!t]
\setlength{\belowcaptionskip}{-0.5cm}
\centering
\includegraphics[width=1\linewidth]{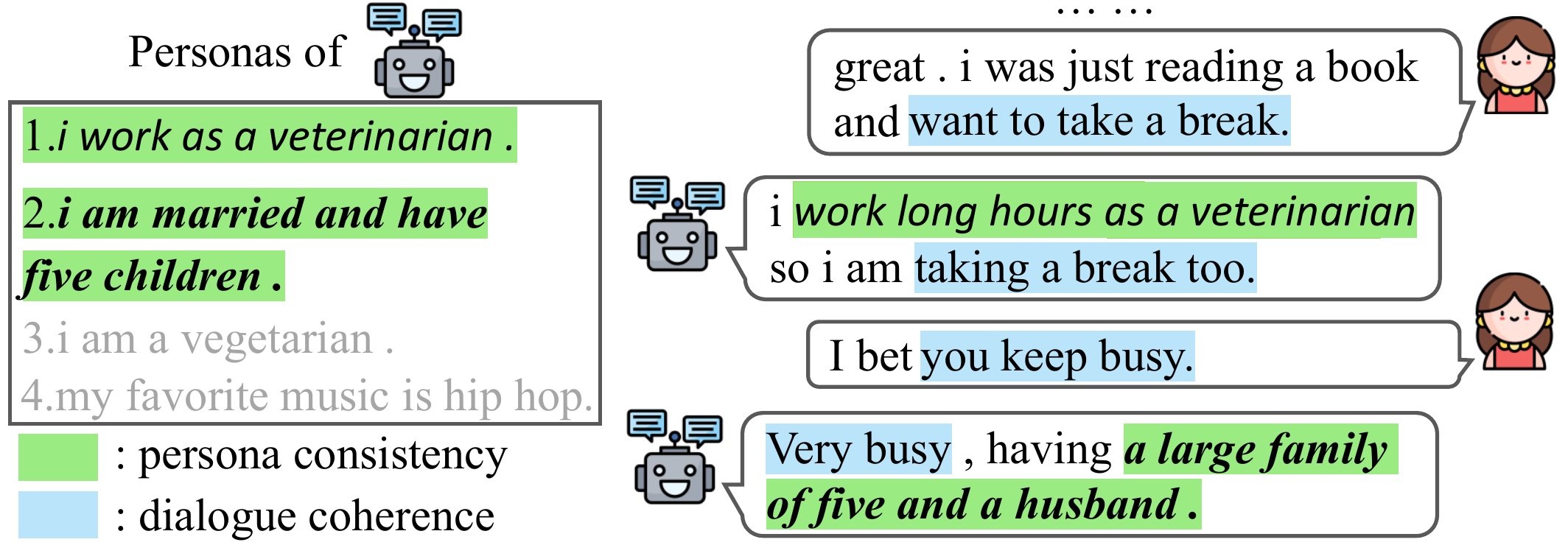}
\caption{\label{fig:persona_example}Each response in a persona-based dialogue is mostly related to one persona sentence and its latest dialogue history utterance. Persona sentences in grey are redundant for all responses. }
%Font styles indicate different persona consistency.}
\end{figure}

%Deep neural dialogue models have shown to be effective when trained on large-scale data, such as Seq2Seq~\cite{seq2seq,seq2seq_attention}, CVAE~\cite{zhao_cvae} and Transformers~\cite{transformer}. The recently proposed models, like OpenAI GPT~\cite{gpt} and GPT2~\cite{gpt2}, also prove their capabilities on dialogue generation tasks~\cite{gpt2_dialogue1,gpt2_dialogue2,transfertransfo}. While most studies focus on single-turn response generation without auxiliary information, efforts are also devoted to personalized dialogue~\cite{personachat}, whose main challenge is that a model needs to condition extra personality texts and dialogue history to generate proper responses that reflect the target persona.
%Deep neural dialogue models have shown to be effective when trained on large-scale data, such as Seq2Seq~\cite{seq2seq_attention}, CVAE~\cite{zhao_cvae} and Transformers~\cite{transformer}. Pretrained language models, like OpenAI GPT~\cite{gpt} and GPT2~\cite{gpt2}, also prove their capabilities on dialogue generation tasks~\cite{gpt2_dialogue1,gpt2_dialogue2}. 
%Recently, 
The ability to generate responses with consistent personas is important towards building intelligent dialogue agents. 
In past years, there has been a growing interest in introducing explicit personas in dialogue generation models~\cite{song_cvae,transfertransfo}. 
A piece of persona text generally consists of profiles and background personal facts. 
%The PersonaChat~\cite{personachat} dataset is a crowd-sourced dataset with data samples covering rich persona features, and 
A clipped persona-based dialogue from the PersonaChat~\cite{personachat} dataset is shown in Figure~\ref{fig:persona_example}, which covers rich persona features.
For a persona-based dialogue generation model, generated responses need to be relevant to the dialogue context as well as consistent with personas.

%Previous work has shown promising results on this task~\cite{song_cvae,transfertransfo}. 
Most existing generation models for this task rely heavily on training with sufficient persona-based dialogues.
However, available data are limited due to their expensive collection costs. 
Take the PersonaChat as an example, two crowd-sourced annotators are hired to play the part of a provided persona and converse naturally with each other.
In total, about 162 thousand dialogue utterances
%~\footnote{Taking the annotations from one annotator as the training target while the other one as the context to avoid repetition.} 
are collected with less than 5 thousand unique persona profiles.
%In comparison, 
Compared with conventional dialogue datasets such as OpenSubtitles~\cite{opensubtitles} and Weibo~\cite{shang2015neural} with millions of utterances, persona-based dialogue datasets are relatively small.

%We now analyze the training difficulty of a persona-based model with current personalized data.
%The first challenge originates from the data format of each training sample. A training sample for persona-based dialogue model involves not only possibly multiple dialogue history utterances but also auxiliary persona sentences.
%On one hand, we compare the data and model for the most simple single-turn short text conversation generation task. 
%A Transformer encoder-decoder model~\cite{transformer} may need millions of data~\cite{opensubtitles,weibo} to perform well on this task. As can be seen, the data scale of PersonaChat is far from enough for a model with similar parameter scales to fit well.

Besides the limited data scale, another data issue we want to point out is that a persona-based dialogue is more complex to learn with, in comparison with conventional dialogues. 
Recall that a persona-based dialogue involves not only multiple dialogue utterances, but also auxiliary persona sentences.
\citet{dialogue_nli} showed that not all responses in the PersonaChat dataset are consistent with the provided personas. This makes it difficult for a model to capture a reliable mapping from training data.
Supposing we apply a similar dialogue model
%Transformer encoder-decoder model~\cite{transformer} 
as in conventional dialogue generation tasks with a comparable parameter size, we should expect more data would be necessary to train a robust model on the more difficult data setting. 
Moreover, it may be difficult to use existing data augmentation methods~\cite{augmentation_cvae,niu_da} to automatically construct such complex persona-based dialogue data. 
%\moe{
%Recent 
%data 
%manipulation methods such as
%data augmentation methods~\cite{back_translation,hou_da_dialogue,guo_mixup_sentence} have shown promising improvements on conventional dialogue generation tasks~\cite{augmentation_cvae,niu_da}. 
%They are often independent of generation models by manipulating the training data only.}
%However, it may be difficult to augment persona-based dialogue data in such a complex format reliably. 
For example, if we apply back translation~\cite{back_translation} to every sentence in persona-based samples, the augmented ones may not maintain the coherence between the dialogue history and the response as well as the consistency between the persona and the response simultaneously.
%On the most simple single-round short text conversation tasks~\cite{}, a Transformer encoder-decoder model~\cite{transformer} often need much more than billions of data to perform well~\cite{}.
%Thus, a similar model structure with comparable parameter sizes 
%Hence, it is essential to study how to train robust and consistent persona-based dialogue generation models with limited personalized dialogues.
%shows a clipped personalized dialogue from PersonaChat~\cite{personachat}. 
%An interlocutor is explicitly described using several persona texts, which makes it harder to generate desired responses as additional personalities need to be conditioned.

%powerful model encounters difficulty in learning on current data 
%when concatenating all these texts as a single long sequence input. The uncertain consistency relationship between the response and each persona text, as well as the large input length, makes it hard for a model to pose enough attention to proper texts. 
% \moe{why? too long, or redundant, or noise, or model limitation?}
% Re: we provide explanations in the former and later sentence
% Moreover, there exist training data that have no relation with any provided profile sentence.
%And the noisy responses with uncertain relationships between personas will further affect model as not all personas are necessary for current generating. 
%However, as noted before, such complex data are very expensive to collect. The PersonaChat dataset only contains 65.7k training samples (one speaker) with only 4.7k unique persona profiles. 
A few studies have been conducted to alleviate the above data issues by finetuning existing pretrained models such as GPT~\cite{transfertransfo,multi_gpt} or BERT~\cite{lin2021adapter,song-etal-2021-bob}. 
They often stick to a certain pretrained model.
Sophisticated finetuning strategies, including proper network modifications and loss functions, are required to get satisfactory performance, making them not useful across different pretrained models. Moreover, they do not address the data difficulty issue explicitly. Most of them simply concatenate all persona and dialogue history sentences into a single input sequence for finetuning, and rely on the ability of the pretrained model to fast adapt to the target data domain.
Hence, we want to design a model-agnostic method to address both the data scale and data difficulty issue, which can be packed with any base model, either trained from scratch or finetuned from a pretrained model.

In this work, we propose a data manipulation method for persona-based dialogue data, which is model-agnostic to be packed with any base model to improve their robustness and consistency.
Our method includes three operations on data, namely \textbf{D}$^3$, in sequence:
(i) \textbf{D}ata distillation: original training samples are simplified into contain only useful and less redundant persona sentences and dialogue utterances, which are expected to be fitted more easily;
(ii) \textbf{D}ata diversification: with the easier distilled samples, we can also perform data augmentation more reliably.  We design various methods to edit new personas, and then align them with new and consistent responses to improve data diversity;
(iii) \textbf{D}ata curriculum: with both augmented distilled and original data at hand, we arrange them into a data curriculum for model learning~\cite{curriculum_learning}, where the base model is trained on the easier augmented distilled data and then the harder original data.
%Our \textbf{D}ata \textbf{D}istillation and \textbf{D}iversification method (\textbf{D$^3$}) is model-agnostic and thus can be packed with any persona-based dialogue generation model. 
To validate the effectiveness of our method, we perform experiments on two strong base dialogue models, Transformer-based encoder-decoder and GPT2. 
%It is also easy to be extended to other models.
%, compared to previous dialogue augmentation methods. Further analyses illustrate the contributions of each part and how our method affects the training.

%Our contributions can be summarized as follows:
% \footnote{***rewrite by urself}
%\begin{itemize}[wide=0\parindent, noitemsep, topsep=0]
%    \item We distill original training data to get simplified persona-consistent samples as an easy data curriculum, helping the model training more effectively.
%    \item We further diversify the distilled data via editing new personas and constructing corresponding aligned responses with quality filtering.
%    \item Extensive experiments and analysis are conducted to demonstrate how \textbf{D$^3$} affects the model.
    % \item We propose a data augmentation technique for personalized dialogue generation.
    % \item Distilled training samples containing consistent persona-response pair are used to constitute an easier learning curriculum before the raw data to train models more efficiently.
    % \item We further diversify the distilled data using pseudo personas and consistent responses after filtering to improve the generation quality.
%\end{itemize}

\section{Related Work}

\paragraph{Persona-based dialogue generation} 
It sees growing interest in recent years, thanks to the released benchmark datasets such as PersonaChat/ ConvAI2~\cite{personachat,convai2}.
Previous works mostly focus on modifying dialogue models to condition auxiliary persona information, including extra persona embedding\cite{li_persona}, profile memory~\cite{personachat}, copying from personas~\cite{deepcopy}, CVAE with persona information~\cite{song_cvae}, and using meta-learning to augment low-resource personas~\cite{tian2021social}.

Recent works try to adopt large-scale pretrained models on this task. 
GPT/GPT2~\cite{gpt,gpt2} are chosen the most often and shown to improve the generation quality with different finetuning strategies~\cite{transfertransfo, multi_gpt,multi-gpt2}. 
Some leverage BERT~\cite{bert} as backbones~\cite{lin2021adapter,song-etal-2021-bob}. 
Other pretrained models also demonstrate their effectiveness~\cite{lin2021adapter}. 
The aforementioned methods often need proper network modifications and finetuning loss functions in order to get satisfactory performance. 
It is hard to transfer them to be useful across different pretrained models. 
Moreover, most of them simply concatenate persona texts and dialogue history together as a single input sequence~\cite{transfertransfo,blender}, highly depending on the ability of the pretrained model to fast adapt to the target data domain. 
%they do not address the data difficulty issue explicitly, highly depending on the ability of the pretrained model to fast adapt on the target data domain.
% 
%However, state-of-the-art results are still far from satisfactory.
%But how to generate responses with a rich and accurate personality remains a challenge.

\paragraph{Text data manipulation} Various data augmentation methods have been widely used in many NLP tasks~\cite{back_translation,hou_da_dialogue,guo_mixup_sentence,min_da}, which are also effective to boost the performance of dialogue models. 
New generated dialogue utterances~\cite{augmentation_cvae,niu_da} and retrieval results~~\cite{dialogue_distillation} can be used to augment the training data.
However, all previous work only studies the pairwise relationship between a query and a response to design the augmentation techniques, which are not applicable to involving auxiliary information, such as personas, simultaneously.
 
Besides data augmentation, there are other ways to manipulate dialogue data to improve model learning. For example, a few approaches filter uninformative or noisy samples to enhance data quality~\cite{augmentation_filter, akama_filter}. \citet{augmentation_manipulation} combine data augmentation and re-weighting to make models learn more effectively. \citet{tian2019augment} utilize learnable memory based on dialogue clusters to enhance the model.

\begin{figure*}[!t]
\setlength{\abovecaptionskip}{0.2cm}
\setlength{\belowcaptionskip}{-0.2cm}
\centering
\includegraphics[width=0.94\textwidth]{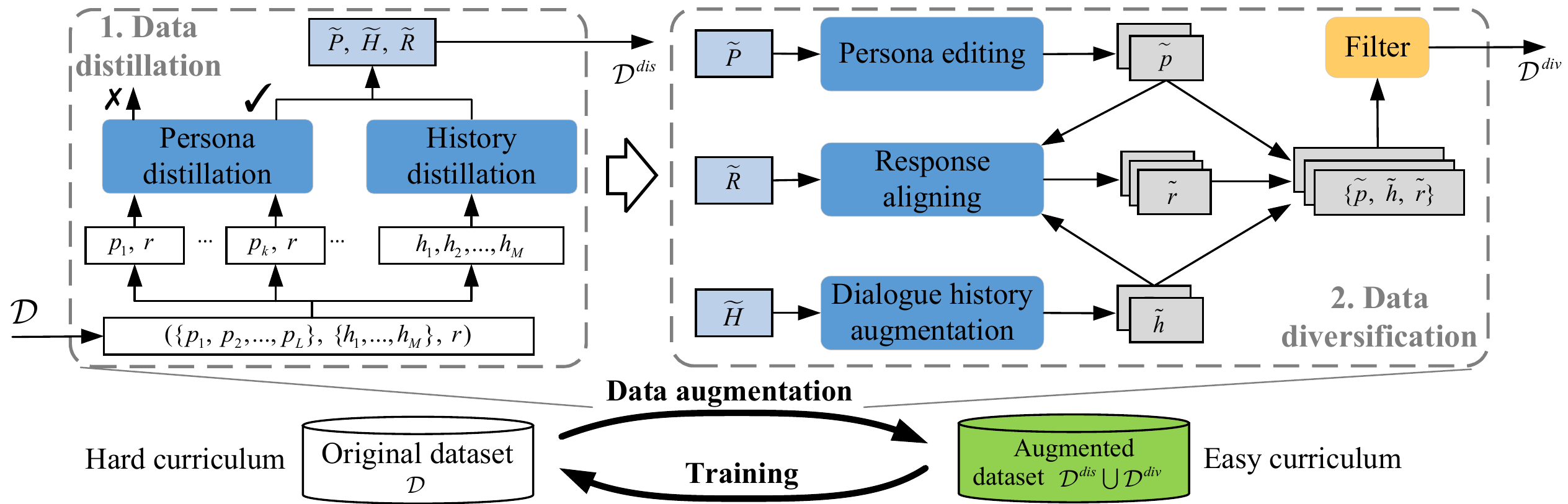}
\caption{The framework of our data manipulation method \textbf{D$^3$}. It obtains the augmented dataset $\mathcal{D}^a=\mathcal{D}^{dis}\cup\mathcal{D}^{div}$ from the original dataset $\mathcal{D}$ through data distillation and data diversification. Curriculum strategy is used to train a model by first learning on the easy augmented data $\mathcal{D}^a$and then on the hard original training data $\mathcal{D}$. }
\label{fig:framework}
\end{figure*}

\paragraph{Curriculum learning} ~\citet{curriculum_learning} examine the benefits of training models using various curricula successively from easy to hard. It has been applied to many NLP tasks such as machine translation~\cite{curriculum_mt}, reading comprehension~\cite{rc_curriculum} and language understanding~\cite{nlu_curriculum}. \citet{dialogue_curriculum} adopt the idea in open-domain dialogue generation, where curriculum plausibility is determined by the response properties, including coherence and diversity. Our work is different in that we introduce new distilled data regarding as a curriculum.
% \textcolor{blue}{Our work is different in that xxx.}

%%%%%%%%%%%%%%%%%%%%%%%%%%%%%%%%%%%%%%%%%%%%%%%%%%%%%%%%%%%%%%%%%
\section{Our Data Manipulation Method}

We first formally define a persona-based training sample. It consists of $L$ persona description sentences $P=\{p_1,p_2,..,p_L\}$, $M$ dialogue history utterances $H=\{h_1,h_2,..,$ $h_M\}$, and a gold response $R$. 
The given training dataset is denoted as $\mathcal{D}=\{(P,H,R)\}$. Note that $L$ and $M$ in different training samples can be different. 
%For an input containing $P$ and $H$ from $\mathcal{D}$, 
A dialogue model needs to generate a response $\hat{R}$, which is coherent with the dialogue history $H$ and consistent with  persona information in $P$.
%Taking PersonaChat~\cite{personachat} as an example, $L$ ranges from 4 to 6, persona texts are simple statements, e.g., ``I favorite music is country music'' or ``I work in sales''.

Our proposed data manipulation method \textbf{D}$^3$ is model-agnostic. For any dialogue model, we will not change the model itself, but only manipulate its training data.
We develop three data manipulation operations in sequel, former two for augmentation and the last one eases training, shown in Figure~\ref{fig:framework}: 
\begin{enumerate}[wide=0pt,noitemsep, topsep=0pt]
    \item \textbf{D}ata distillation. We construct simple persona-consistent data $\mathcal{D}^{dis}=\{(\widetilde{P}, \widetilde{H}, \widetilde{R})\}$ by removing redundant information in $P$ and $H$;
    \item \textbf{D}ata diversification. Due to the limited amount of distilled samples, we design various methods to increase the data variety and scale, and obtain the diversified data $\mathcal{D}^{div}=\{({\widetilde{p}}, {\widetilde{h}}, {\widetilde{r}})\}$;
    \item \textbf{D}ata curriculum. We combine $\mathcal{D}^{dis}$ and $\mathcal{D}^{div}$ as the augmented dataset $\mathcal{D}^a$. A curriculum strategy is defined to train the model with the easier distilled samples in $\mathcal{D}^a$ first and then the original ones in $\mathcal{D}$.
\end{enumerate}

\subsection{Data Distillation}
\label{sec:data_distillation}
Before introducing our distillation method, we discuss the difficulty of training a model with the original training samples in detail.
The dependency of a response on the given persona fluctuates between different parts of the persona sentences. As shown in Figure~\ref{fig:persona_example}, most responses only correspond to one persona sentence. The remaining persona information is mostly redundant, and may confuse the model to attend on useful persona information. 
Similarly, we notice that models tend to attend more on the last few utterances of $H$ rather than the historical ones. We find that by using a Transformer encoder-decoder model, the attention weights of the last Transformer layer on the last utterance is 45\% higher than the average on the other utterances. See  Appendix~\ref{app:attention_history} for the experiment and results. This observation is also consistent with previous studies on multi-turn context understanding~\cite{context_length,dialogue_length}.
% \footnote{***do  u have any statistical number here?}

A few previous works have demonstrated that attention-based models will be distracted by noisy attended information, and accurate attention supervisions can be very beneficial~\cite{liu2016neural,hsu_attention}.
Inspired by them, we mimic a ``hard'' attention supervision between the response and useful persona/dialogue history by directly removing redundant tokens in the attended sequences. Therefore, different from previous work that modify the model to inject attention supervisions, our method only manipulates data.
% one main challenge of this task is that multiple persona profile may disturb the model to correctly learn attention on the most suitable persona text. And some general responses irrelevant to any persona will further confuse the model. 
%Therefore, 
%We propose to distill an original sample into a new one such that the target response is now highly determined on the provided persona and dialogue history.
%
%This is also connected with previous work, in which models benefit from data with supervised attention~\cite{liu2016neural,hsu_attention}.
%Here, we also mimic to output ``hard'' attention alignment between the response and useful persona texts/dialogue history by simply removing the unaligned information.
%Unlike previous work that inject supervision by modifying the model, our method only manipulates data. 
%In the following, 
%We distill an original sample into a new one by separately removing redundant sentences in its persona and dialogue history, respectively.

%And we further simplify them to remain a shorter dialogue history so as to reduce the input sequence length. 
%Our method only makes manipulations on data level, and model learning can benefit from a preferable data without any extra attention supervision like former work~\cite{hsu_attention}.
% Differs from former works who offer extra attention supervision on desired positions~\cite{hsu_attention}, 

\paragraph{Persona distillation} 
We aim to determine which persona sentence the current response is consistent with, and thus remove the remaining non-consistent ones. To do so, we associate each persona sentence $p_k$ with the target response $R$ 
%to form a series sentence pairs $\{(p_1, R), (p_2, R),...,(p_L, R)\}$
, and 
%We formulate this problem as 
determine the consistency between each $p_k$ and $R$.
Following previous work~\cite{dialogue_nli}, we cast it as a natural language inference (NLI) problem.
%in which a model needs to determine whether a sentence $R$ entails the other sentence $p_k$. 
If $R$ entails $p_k$ , it is considered to be consistent with $p_k$, otherwise irrelevant to $p_k$. 
A trained RoBERTa~\cite{roberta} model is used here as the NLI model, with an accuracy of 90.8\% on the DialogueNLI dev set provided in~\citet{dialogue_nli}. Details are provided in Appendix~\ref{app:detail_distillation}. 
%Given a sentence pair $(P_k, R)$, we apply
%\begin{equation}
%    p = Softmax(NLI(P_k, R)),
%\end{equation}
%where $p$ is the probability for two texts being entailed.
%If $p_e$ is the largest then $(P_k, R)$ is our distillation target. We also set a threshold $\tau$ and ensure that $p_e \geq \tau$ to filter low-confidence pairs.
% If $p_e$ is larger than other two values, then this pair should be our distillation target and $R$ is consistent with $P_k$. In addition, considering the error rate of NLI model, we filter low-confident samples and set a threshold $\tau>0$ that only sentence pairs with $p_e \geq \tau$ are selected.

\paragraph{Dialogue history distillation} 
%The raw dialogue history $H$ is usually long and most information is redundant for a coherent $\hat R$, which may cause extra distraction. Besides, 
We can adopt a trained attention-based model to determine useful context sentences. 
%However, we do not want our method dependent on any dialogue model.
For simplicity, we could also keep only the most useful last utterance $H_M$ in a distilled sample (as suggested by our preliminary experiments discussed in the beginning of this section). In our experiments in \secref{sec:expt_main}, we find that using the last utterance is enough for our method to work well.
 %, which should ease the model learning while also guaranteeing the generation coherence. 

A distilled sample $(\widetilde{P}, \widetilde{H}, \widetilde{R})$ is ready to be constructed now. Here, $\widetilde{P}$ and $\widetilde{H}$ both contain only one sentence. $\widetilde{P}$ is any $p_k$ that entails $R$, and $\widetilde{H}$ is the last utterance in the dialogue history, and $\widetilde{R}=R$. 
Such samples form the distilled dataset $\mathcal{D}^{dis}$. 
Note that an original sample in $\mathcal{D}$ may result in none, one, or multiple distilled samples, as $R$ may entail none, one, or multiple persona sentences.

\subsection{Data Diversification}

Distilled samples should ease model training as their responses are highly dependent on their $\widetilde P$ and $\widetilde H$.
However, samples in $\mathcal{D}^{dis}$ are limited in terms of both \textbf{scale} (around 40\% of the original data) and \textbf{diversity} (about 4.5k unique persona sentences).
%, which may affect the training efficiency of data-driven models so that personalized dialogue generation. 
Hence, it is necessary to augment $\mathcal{D}^{dis}$. 
%
%Some studies validate the advantage of adding augmented samples on conventional dialogue tasks~\cite{augmentation_cvae,augmentation_manipulation}.
%But these methods only consider the query-response dialogue pairs and cannot handle the more complicated dependency between dialogue and the auxiliary information such as personas.
%
Thanks to the assured relationship between $\widetilde P$/$\widetilde H$ and $R$, we can devise possible methods to diversify distilled samples with more semantically varied samples.
Our data diversification operation contains the following three parts
%: persona editing, dialogue history augmentation, and response 
%aligning 
along with quality filtering, as shown in Figure~\ref{fig:framework}.
%, starting from a distilled sample $(\widetilde{P}, \widetilde{H}, \widetilde{R})$.

\paragraph{Persona editing} 
%To obtain a good persona-consistent dialogue model, 
%Since current $\widetilde R$ reflects persona, 
%It is essential to involve more diverse persona texts in order to learn a robust persona-consistent model. 
We aim to obtain new persona sentences to improve the data scale, and more importantly the persona diversity. Hence, we here consider both token-level and phrase-level editing methods given a persona sentence $\widetilde{P}$:
\begin{itemize} [wide=0pt,noitemsep, topsep=0pt]
    \item Token-level editing: we randomly mask a pre-defined ratio of tokens in $\widetilde{P}$, then use a pretrained BERT~\cite{bert} model to make predictions on the masked positions one by one. 
    %The new tokens will take place of the old ones.
    \item Phrase-level editing: we remove the last few tokens in $\widetilde{P}$  with the removal length determined by a random ratio, and utilize a pretrained GPT2~\cite{gpt2} to rewrite the removal part. 
\end{itemize}
%
% Here, $\rho^t$ is a fixed token-level mask ratio and $\rho^p$ is a variable ratio sampled between $[\rho^p_{min},\rho^p_{max}]$. 
Multiple edited persona sentences can be obtained from one certain $\widetilde P$.
Here, we finetune pretrained models using all persona sentences for a trade-off between semantic diversity and domain similarity.
% in order to generate semantically-diverse persona sentences with higher probabilities.
%Figure~\ref{fig:generate_persona} illustrate an editing case, showing that the new persona texts can effectively increase personality diversity.
%
%Although persona descriptions are usually simple and these approaches can get reasonable results, we filter
To ensure a satisfactory fluency and novelty of an edited persona $\widetilde p$, we rate it via a scoring function:
\begin{equation}
    f = \alpha \cdot \text{PPL}(\widetilde p) + (1 - \alpha) \cdot \text{BS}_f(\widetilde p, \widetilde P).
    \label{eq:1}
\end{equation}
Here, $\text{PPL}$ calculates the normalized perplexity via a GPT2 model to measure its fluency, and the rescaled F1 value of BERTScore ($\text{BS}_f$)~\cite{bert_score} is employed to evaluate the semantic similarity between two sentences. 
Lower values for both functions are preferred, indicating higher fluency or novelty. $\alpha$ is a hyper-parameter. We rank all edited personas originated from the same $\widetilde P$ with the ascending order of their scores in Eq.~\ref{eq:1}, and select the top $N_p$ ones.
%to form the diversified persona set ${\widetilde P}^d=\{{\widetilde p}_i\}_{1 \leq i \leq N_p}$.

\begin{figure}[!t]
    \setlength{\abovecaptionskip}{0.2cm}
    \setlength{\belowcaptionskip}{-0.2cm}
    \centering
    \includegraphics[width=1\linewidth]{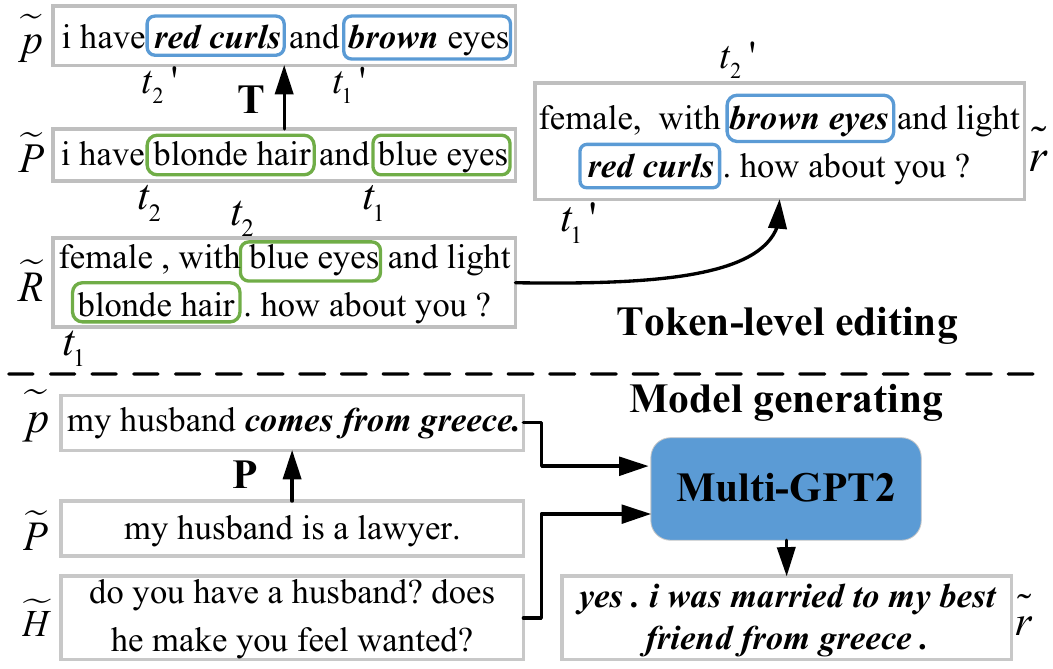}
    \caption{Aligning responses for new personas via token-level editing or model generating. \textbf{T}/\textbf{P}: edit persona in token/phrase level.($t_1$ and $t_2$ are overlapped tokens, $t_1'$ and $t_2'$ are corresponding new edited and aligned tokens.)}
    \label{fig:generate_response}
\end{figure}

\paragraph{Response aligning}
%Corresponding response for a pseudo persona reflecting its characters are essential to form a complete new sample, matching the distilled data. 
Since the semantic meaning of an edited persona sentence obtained above could change, the original response may not be consistent with it. Therefore, we need to get a new aligned response to maintain the persona consistency.
%Such responses are obtained based on the permutations of each text from ${\widetilde P}^a$ and ${\widetilde H}^a$.
Two approaches are utilized to obtain an aligned response ${\widetilde r}$ given an edited persona sentence ${\widetilde p}$ and the corresponding distilled history utterance ${\widetilde H}$:
\begin{itemize} [wide=0\parindent,noitemsep, topsep=0pt]
\item  Token-level editing: We observe that some overlapped tokens can be found between $\widetilde P$ and $\widetilde R$. If an overlapped token $w$ has been changed to a new token $w'$ in the edited persona ${\widetilde p}$, we directly replace $w$ in $\widetilde R$ with $w'$ in the same positions, resulting in an aligned response ${\widetilde r}$. 
An illustration figure can be found in Appendix~\ref{app:data_diversification}.
\item  Model predicting: If no overlapped token can be found, token-level editing will not be applicable. Then we employ a GPT2-based encoder-decoder model~\cite{multi-gpt2} finetuned on the distilled data $\mathcal{D}^{dis}$ to predict responses with the given ${\widetilde p}$ and a dialogue history utterance ${\widetilde H}$. 
%Then we can obtain a series of responses to form diversified samples.
\end{itemize}
Figure~\ref{fig:generate_response} demonstrates the two kinds of approaches.

\paragraph{Dialogue history augmentation}
%The semantic of an edited persona text obtained above could change and, thus the original response may not be consistent with it. Therefore, we need to get a new response to recover the consistency. To ease this operation, here we do not aim to edit dialogue history to promote diversity.
%As we need to extend the data scale but without large semantic divergence on history which may affect our following process,
To further scale up the size of distilled samples, we also manipulate the dialogue history ${\widetilde H}$.
%We can apply a similar method in persona editing to first edit the history utterances and then obtain the new coherent responses. 
Since the diversity scarcity issue is not severe in ${\widetilde H}$, we use a popular sentence-level data augmentation method, back translation (BT)~\cite{back_translation}, to obtain variants of dialogue utterances. We could consider the semantics of the variants are identical. 
Distilled history utterance $\widetilde H$ is translated into an intermediate language, then back into the source language using a couple of existing translation models. The original dialogue history and its $N_h$ variants compose the augmented dialogue history set $\{{\widetilde h}\}$.
%=\widetilde H \cup \{{\widetilde h_j}\}_{1 \leq j \leq N_h}$.

Combining the above three parts together, we now obtain new samples 
%responses ${\widetilde R}^d=\{{\widetilde r}_{ij}^d\}$ by the permutations of each item from 
%${\widetilde P}^d$ and ${\widetilde H}^d$. To ensure the quality of each new sample 
$\{(\widetilde p, \widetilde h, \widetilde r)\}$. We evaluate them with respect to fluency, persona consistency and history coherence:
%, which are all desired generation goals in generation. The score of a new sample ${\widetilde S}^a=({\widetilde P}^a_i, {\widetilde H}^a_j, {\widetilde R}^a_{ij})$
\begin{multline}
     s = \beta \cdot \text{PPL}({\widetilde r}) + \gamma \cdot \text{NLI} ({\widetilde p}, {\widetilde r}) \\
      + (1 - \beta - \gamma)\text{NLI}_c(\widetilde h, \widetilde r),
      \label{eq:filter}
\end{multline}
where $\text{NLI}$ measures the entailment between a persona sentence and the response by the same NLI model in \secref{sec:data_distillation}, and $\text{NLI}_{c}$ evaluates the entailment between a dialogue history utterance and the response using another NLI model~\cite{nli_coherence}(details in Appendix~\ref{app:data_diversification}). $\beta$ and $\gamma$ are hyper-parameters. 
We filter samples below a threshold $T$, and the remaining samples constitute the diversified data set $\mathcal{D}^{div}$. The whole augmented training dataset is the union of $\mathcal{D}^{dis}$ and $\mathcal{D}^{div}$.
The quality of augmented samples is discussed in Appendix~\ref{app:details_experiments}.
% It will be put together with the original distilled samples as the augmented training dataset $\mathcal{D}^a$.

\subsection{Data Curriculum}

%After get $\mathcal{D}^a$, we need to integrate it in training. Although samples in $\mathcal{D}^a$ can benefit model by the determinate persona-response relations and simpler format, the raw data $\mathcal{D}$ cannot be discarded. 
During inference, the model should be capable to handle testing data with multiple persona sentences and dialogue history utterances as the original data. 
% Moreover, we can not obtain its distilled version since the response is not exposed.
Therefore, a model trained using $\mathcal{D}^a$ only is not proper.
We should use both $\mathcal{D}^a$ and $\mathcal{D}$.
%To benefit the model training with the augmented distilled dataset $\mathcal{D}^a$ but not 
Unlike previous studies that treat the original and augmented data equally and mix them directly, we design a curriculum strategy.
%because they are in different information levels. Therefore, inspired by curriculum learning~\cite{curriculum_learning}, 
Considering the different training difficulty of data in $\mathcal{D}^a$ and $\mathcal{D}$, we treat $\mathcal{D}^a$ as an easy curriculum while the original dataset $\mathcal{D}$ as a hard curriculum. 
%Because we remove some distractors in $\mathcal{D}^a$. 
The model is trained on such data curriculum successively until convergence.
% Most related works regard augmented data as parallel of the raw one and mix them together for training. Nevertheless, our augmented data is a refined from raw data that only contain determinate persona-consistent samples in a simplified format. It is easier for models to learn how to utilize persona profiles in responses on such data. Inspired by curriculum learning~\cite{curriculum_learning}, instead of using mixture, we use augmented data as the easy curriculum while the raw data as hard curriculum to train the model successively. We do not discard the raw data as models still need to handle general conditions with no need of indicating persona and augmented data misses this part.

\begin{table}[t!]
\setlength{\tabcolsep}{1.4mm}
\small
\centering
\begin{tabular}{l|c|c|c|c|c}
\toprule
 & $\mathcal{D}$ & $\mathcal{D}^{dis}$ & $\mathcal{D}^{div}$ & $\mathcal{D}^a$ & $\mathcal{D} + \mathcal{D}^a$\\
\midrule
\#sample & 65,719 & 26,693 & 26,700 & 53,393 & 119,112\\
\#persona & 4,710 & 4,522 & 9,788 & 14,310 & 14,498 \\
\#token & 20,467 & 13,420 & 12,794 & 17,835 & 23,269 \\
\bottomrule
\end{tabular}
\caption{\label{tab:statistics} Statistics of samples obtained in each stage.}
\end{table}

\section{Experiments}
\label{sec:expt_main}
To validate the effectiveness of our proposed model-agnostic data manipulation method, we first experiment on two strong persona-based dialogue generation models (Transformer encoder-decoder and GPT2) on the benchmark PersonaChat~\cite{personachat} dataset. Next we conduct a series of analysis to examine the usefulness of different data manipulation operations in our method.
\footnote{Code is available at \url{https://github.com/caoyu-noob/D3}.}
\subsection{Experimental Setup}
\paragraph{Dataset} The PersonaChat~\cite{personachat} data is widely used in this field~\cite{song_cvae,song_edit,transfertransfo,multi_gpt}. 
%Although other datasets may exist, it is the most commonly-accepted one and it is easy to make comparison. 
%It contains 8,939/1,000/968 multi-turn dialogues in Train/Dev/Test set respectively, totally 164,356 utterances. We consider the %\textsc{Self} \textsc{Original} set with fewer samples for a harder setting. 
Each sample has a dialogue history $H$ with no more than 15 utterances ($M\leq15$) and a persona $P$ with between 4 and 6 sentences ($4\leq L\leq6$). 
% We remain the last 5 utterances in $H$ as the dialogue history.
%For \textbf{D$^3$}, we set the distillation threshold $\tau=0.99$, the edited persona number $N_p=5$. A suitable filtering threshold $T$ extends distilled data into about 200\% of its original size in diversification
Numbers of samples, unique persona sentences, and tokens in each stage of our method are listed in Table~\ref{tab:statistics}. 
% \footnote{\#persona is the number of unique sentences or persona?}

\begin{table*}[t!]
\setlength{\tabcolsep}{1.15mm}
\small
\centering
\begin{tabular}{l|cccc|cccccc|c|ccc}
\toprule
\textbf{Model} & PPL & BLEU & NIST-4 & BS$_f$ & Ent-1 & Ent-2 & Ent-3 & Dis-1 & Dis-2 & Dis-3 & \textbf{C} & Flu. & Coh. & Pcon. \\
\midrule
Human & - & - & - & - & 5.680 & 8.913 & 10.27 & 5.259 & 34.90 & 66.37 & 0.472 & 2.625 & 2.451 & 0.531 \\
\midrule
\textsc{Trans} & 38.28 & 3.140 & 1.148 & \cellcolor{gray!20}0.1486 & 4.046 & 5.484 & 6.262 & 1.609 & 6.298 & 11.71 & \cellcolor{gray!20}0.235 & \cellcolor{gray!20}2.303 & \cellcolor{gray!20}2.038 & \cellcolor{gray!20}0.304 \\
\textsc{Trans-BT} & 37.92 & \underline{3.315} & 1.082 & \cellcolor{gray!20}0.1527 & \underline{4.274} & 5.905 & 6.752 & 1.760 & 7.108 & 13.39 & \cellcolor{gray!20}0.289 & \cellcolor{gray!20}\underline{2.337} & \underline{2.142} & \cellcolor{gray!20}0.350 \\
\textsc{Trans-CVAE} & \underline{37.61} & 3.312 & \underline{1.191} & \cellcolor{gray!20}0.1533 & 3.974 & 5.451 & 6.267 & 1.459 & 5.795 & 11.16 & \cellcolor{gray!20}0.260 & \cellcolor{gray!20}2.333 & \cellcolor{gray!20}2.111 & \cellcolor{gray!20}0.335 \\
\textsc{Trans-filter} & 38.99 & 2.946 & 1.101 & \cellcolor{gray!20}\underline{0.1563} & \bf 4.283 & \underline{6.033} & \underline{7.088} & \underline{1.796} & \underline{7.696} & \underline{14.06} & \cellcolor{gray!20}\underline{0.446} & \cellcolor{gray!20}2.318 & \cellcolor{gray!20}2.088 & \cellcolor{gray!20}\underline{0.492} \\
\bf \textsc{Trans-\textbf{D$^3$}} & \bf 37.30 & \bf 3.358 & \bf 1.206 & \bf 0.1574 & 4.223 & \bf 6.165 & \bf 7.298 & \bf 1.826 & \bf 7.923 & \bf 14.42 & \bf 0.485 & \bf 2.397 & \bf 2.172 & \bf 0.513 \\
% \textsc{Trans-\textbf{D$^3$}*} & \underline{37.67} & 3.259 & 1.185 & 0.1554 & 4.197 & \underline{6.095} & \underline{7.232} & 1.794 & \underline{7.835} & \underline{14.27} & 0.439 & \underline{2.378} &  \underline{2.164} & 0.481 \\
\midrule
\textsc{GPT2} & 17.63 & 3.761 & 1.278 & \cellcolor{gray!20}0.1693 & 4.485 & 6.187 & 7.029 & 2.011 & 8.260 & 15.03 & \cellcolor{gray!20}0.518 & 2.508 & 2.243 & 0.508 \\
\textsc{GPT2-BT} & 16.96 & \underline{3.943} & 1.348 & \cellcolor{gray!20}0.1663 & 4.547 & 6.248 & 7.089 & 1.947 & 8.113 & 14.94 & \cellcolor{gray!20}0.509 & \cellcolor{gray!20}2.488 & \bf 2.259 & \cellcolor{gray!20}0.454 \\
\textsc{GPT2-CVAE} & 17.16 & 3.339 & \underline{1.360} & \cellcolor{gray!20}0.1592 & 4.245 & 5.691 & 6.490 & 1.748 & 6.799 & 12.19 & \cellcolor{gray!20}0.484 & \cellcolor{gray!20}2.358 & \cellcolor{gray!20}2.150 & \cellcolor{gray!20}0.426 \\
\textsc{GPT2-filter} & \underline{16.90} & 3.734 & 1.337 & \cellcolor{gray!20}0.1788 & \underline{4.570} & \underline{6.352} & \underline{7.263} & \underline{2.148} & \underline{9.031} & \underline{16.52} & \bf 0.571 & \underline{2.527} & 2.233 & \underline{0.537} \\
\bf \textsc{GPT2-\textbf{D$^3$}} & \bf 15.69 & \bf 4.184 & \bf 1.429 & \bf 0.1835 & \bf 4.614 & \bf 6.426 & \bf 7.321 & \bf 2.267 & \bf 9.803 & \bf 18.20 & \underline{0.557} & \bf 2.532 & \underline{2.255} & \bf 0.548 \\
% \textsc{GPT2-\textbf{D$^3$}*} & \underline{15.77} & \underline{4.082} & \underline{1.388} & \underline{0.1809} & \underline{4.611} & \underline{6.408} & \underline{7.312} & \underline{2.209} & \underline{9.657} & \underline{17.91} & 0.536 & 2.525 & 2.249 & 0.527 \\
\bottomrule
\end{tabular}
\caption{\label{tab:main_result}Results of all compared data manipulation methods on two base models. BLEU and Dist-n are in \%. Best results are in bold, and second best are underlined. 
% \textbf{D$^3$}* means using an NLI model trained under a few-show setting (200 samples) in the data distillation.
Shaded numbers indicate our \textsc{D$^3$} is significantly better than this method on human evaluation, C-score and BS$_f$, accoding to our significance T-test where $p > 0.05$.}
\end{table*}

\paragraph{Base models} Two dialogue model architectures are considered:
\begin{itemize}[wide=0\parindent,noitemsep, topsep=0pt]
    \item \textsc{Transformer}~\cite{transformer}: an encoder-decoder architecture using Transformer as the backbone with pointer generator~\cite{pointer} integrated;
    \item \textsc{GPT2}: one of the most powerful pretrained models on this task~\cite{transfertransfo,multi_gpt,multi-gpt2}.
\end{itemize}
\textsc{Transformer} is trained from scratch, and \textsc{GPT2} is finetuned. For both models, we construct training data by concatenating persona and dialogue history as a single input sequence, in which special symbols and token type embeddings are involved to distinguish between them. The negative log-likelihood loss is used to train models using Adam optimizer~\cite{adam}.

\paragraph{Compared methods} We pack two base models with our method \textbf{D$^3$} and other data manipulation approaches for comparison: 
\begin{itemize}[wide=0\parindent,noitemsep, topsep=0pt]
\item \textsc{Back Translation (BT)}~\cite{back_translation}:
we perform BT on all sentences in a training sample, including the persona sentences and dialogue utterances, and train the model with the augmented and original data jointly;
%it diversify the corpus by translating the source language into a intermediate language and then translate back, obtaining a text that  differs from the original one;
\item \textsc{CVAE}~\cite{augmentation_cvae}:  a CVAE-based generation model is trained on the original data and then used to generate new responses via sampling with different latent codes. Since it can only handle pairwise data, we concatenate all input sentences as a single input sequence in this method;
\item \textsc{Entropy Filter (filter)}~\cite{augmentation_filter}: it removes generic responses according to the entropy, which is calculated using the dialogue history and the response without using the persona. 
\end{itemize}
%As our base models can achieve competing performance among existing works, we do not focus on comparing with other network architectures.
The detailed configurations of each method are given in Appendix~\ref{app:details_experiments}.

\paragraph{Automatic metrics} We adopt multiple widely used metrics to measure the response quality, including Perplexity (PPL), BLEU~\cite{bleu}, NIST-4~\cite{nist} and BERTScore~\cite{bert_score}. We use the same BS$_f$ in Eq.~\ref{eq:1} for BERTScore. To evaluate the response diversity, we use Distinct-n~\cite{li_diversity} (Dist, n=1,2,3) which is the ratio of unique n-grams among the corpus, and Entropy-n~\cite{zhang_entropy} (Ent, n=1,2,3) that is the entropy obtained via the n-gram distribution in a sentence. Moreover, C-score~\cite{cscore} (\textbf{C}) is involved, where we follow the default setting and use the output of an NLI model trained on the DialogueNLI dataset~\cite{dialogue_nli} to indicate the consistency between a response and persona sentences. 
% \footnote{***which nli model used here?}

\paragraph{Human evaluation} 
%Human evaluation is also involved in our experiments. 
We randomly selected 200 samples from the test set for human evaluations. Five professional annotators from a third-party company were asked to rate the responses from three aspects: 1) Fluency (Flu.); 2) Coherence (Coh.) with the dialogue history, 3) Persona consistency (Pcon.). The scores for the first two aspects have three scales, in which 1/2/3 indicates unacceptable/moderate/satisfactory respectively. The last one is binary, where 1 means the response is consistent with at least one persona sentence in the sample and 0 otherwise. 
The agreement rate from raters is 97.5\%, 89.5\%, 100\% @3 (at least 3 of them reach an agreement) in the these aspects, indicating the validity of scores. The instruction of human evaluation is given in Appendix~\ref{app:details_experiments}.
% \footnote{***what does @3 here mean?}
% I have added an explanation

\subsection{Results}

Table~\ref{tab:main_result} reports the results on two based models trained with the use of various compared data manipulation methods. 
T-test is conducted between our \textsc{D$^3$} and other compared methods on each base model for metrics including BS$_f$, C-score and three human evaluation metrics. Other automatic metrics have similar results or are not applicable such as Distinct-n. %Results are indicated by cell color in Table~\ref{tab:main_result}. 
Details of the significant tests are given in Appendix~\ref{app:statistics_results}.

\begin{table*}[t!]
\setlength{\tabcolsep}{1.4mm}
\small
\centering
\begin{tabular}{l|cccc|cccccc|c}
\toprule
 & PPL & BLEU & NIST-4 & BS$_f$ & Ent-1 & Ent-2 & Ent-3 & Dis-1 & Dis-2 & Dis-3 & \textbf{C} \\
\midrule
\textsc{Trans} & 38.28 & 3.140 & 1.148 & 0.1486 & 4.046 & 5.484 & 6.262 & 1.609 & 6.298 & 11.71 & 0.235 \\
\textsc{Trans-\textbf{D$^3$}} & 37.30 & 3.358 & 1.206 & 0.1574 & 4.223 & 6.165 & 7.298 & 1.826 & 7.923 & 14.42 & 0.485 \\
\textsc{Trans-\textbf{D$^3$}*} & 37.67 & 3.259 & 1.185 & 0.1554 & 4.197 & 6.095 & 7.232 & 1.794 & 7.835 & 14.27 & 0.439 \\
\midrule
% \quad \textit{only augmented data} & 126.3 & 1.603 & 0.891 & .8456 & \textit{4.368} & \textit{6.390} & \textit{7.404} & 1.762 & 7.207 & 12.97 & \textit{0.942} \\
\quad \textit{w/o diversification} & 37.90 & 3.159 & 1.105 & 0.1511 & 4.051 & 5.664 & 6.533 & 1.570 & 6.992 & 13.42 & 0.454 \\
\quad\quad \textit{w/o distillation} & 38.25 & 3.105 & 1.126 & 0.1499 & 4.026 & 5.459 & 6.290 & 1.495 & 6.131 & 11.76 & 0.352 \\
\quad \textit{only distillation} & 104.8 & 1.509 & 0.939 & 0.1059 & 4.002 & 5.398 & 6.265 & 1.279 & 4.630 & 8.505 & 0.637 \\
%\midrule
%GPT2-D2D & 15.69 & 4.184 & 1.429 & .8622 & 4.614 & 6.426 & 7.321 & 2.179 & 9.458 & 17.72 & 0.557 \\
%\quad \textit{only augmented data} & 33.01 & 2.540 & 1.097 & .8487 & 4.438 & 6.331 & \textit{7.335} & 1.313 & 4.819 & 8.714 & \textit{1.148} \\
%\quad \textit{w/o diversification} & 15.91 & 4.118 & 1.421 & .8619 & 4.517 & 6.229 & 7.012 & 2.000 & 8.688 & 16.29 & 0.521 \\
%\quad \textit{w/o distillation} & 16.73 & 4.021 & 1.406 & .8613 & \textit{4.644} & \textit{6.481} & 7.215 & \textit{2.239} & 9.283 & 17.45 & 0.503 \\
\midrule
\quad \textit{w/o persona} editing & 37.96 & 3.284 & 1.136 & 0.1535 & 4.171 & 5.686 & 6.517 & 1.608 & 6.599 & 12.62 & 0.422 \\
\quad \textit{w/o history augmentation} & 38.10 & 3.291 & 1.222 & 0.1550 & 4.150 & 5.759 & 6.560 & 1.608 & 6.493 & 12.52 & 0.461 \\
%\quad w/o persona filter & 37.74 & 3.253 & 1.160 & .8569 & 4.272 & 5.921 & 6.971 & 1.627 & 7.463 & 13.91 & 0.446 \\
\quad \textit{w/o response filter} & 38.21 & 3.106 & 1.087 & 0.1503 & 4.207 & 5.841 & 7.080 & 1.592 & 6.991 & 12.98 & 0.399 \\
\bottomrule
\end{tabular}
\caption{\label{tab:analysis}Automatic evaluation results with variant in data distillation (middle), and diversification (bottom), compared with our full method (top) on \textsc{Transformer}. \textbf{D$^3$}* means using an NLI model trained under a few-show setting (200 labelled samples) in the data distillation. }
\end{table*}

%Fo, compared to training with the original dataset or other data manipulation methods, our method obtains the best persona consistency.
%, e.g., 70\% higher than the base Transformer. 
%Our method shows less improvement on GPT2 than Transformer, but many former data-level methods even fail on GPT2. 
On \textsc{Transformer}, all methods achieve improvements on most metrics compared with training with the original dataset. Our method yields the best performance except for Ent-1.
On \textsc{GPT2}, many methods fail to improve the various metrics consistently. For example, on the persona consistency (Pcon.), only \textsc{entropy filter} and our method can get higher scores than training with the original dataset.
The reason is that the data scarcity issue is less severe with a pretrained model, and it is more important to address the data diversity issue.
%The reason is that Transformer is an end-to-end model while GPT2 is pre-trained on a huge corpus and data issues may have a less significant impact.
In our method, the augmented distilled samples are encouraged to have different semantics with the original ones and improve the data diversity, and thus continue to get improvements on the strong pretrained GPT2.

\subsection{\label{sec:more_analysis}More Analysis}
We further analyze the contributions made by different data manipulation operations in our method by answering the following three questions:
\begin{enumerate}[wide=0\parindent,noitemsep, topsep=0pt]
\item Is there a need to construct simple data $\mathcal{D}^{dis}$ as in data distillation?
\item Can data diversification effectively obtain diverse distilled data?
\item Does the curriculum strategy better exploit the augmented data and help model training?
\end{enumerate}
We use results on \textsc{Transformer} here for discussion in the following part. Refer to Appendix~\ref{app:gpt2_ablation} for extensive results on \textsc{GPT2} model. 
%We use automatic metrics here. Despite they are not so reliable among different model architectures, they can basically reflect the performance gaps under the same architecture based on our observation in Table~\ref{tab:main_result}.

\paragraph{Analysis of data distillation}
To examine the effectiveness of data distillation, we need to neutralize the influence of data diversification as it is only applicable to distilled data. Following variants of our \textbf{D$^3$} are considered: 
1) \textit{w/o diversification}: only using distilled data $\mathcal{D}^{dis}$ in the easy curriculum;
2) \textit{w/o distillation}: based on 1), we recover samples in $\mathcal{D}^{dis}$ into their original format, which means all their persona sentences and history utterances are included; 
3) \textit{only distillation}: only $\mathcal{D}^{dis}$ is used in training without using the original data in $\mathcal{D}$.

Results of these variants are shown in the middle of Table~\ref{tab:analysis}.
Obviously, removing data diversification decreases the performance in all aspects as the model has less training data. 
If we further remove data distillation and use the same amount of data in their original formats, the model performs even worse, especially on the C-score. 
This validates the effectiveness of data distillation in our method. 
%Although $\mathcal{D}^{dis}$ only contains responses that are consistent with at least one persona which should be easier for model learning than the original data, 
However, it is not proper to completely rely on distilled data. 
From the results of only using distilled data in training,  our method improves the C-score, yet significantly degenerates in other aspects.
The reason is that the relationship between persona/dialogue history and the response has changed from the original data to their distilled ones. Thus a model trained with distilled data should serve as a warm start to learn the original data, but not to replace the original data.

We also test the robustness of our data distillation method by using an NLI model trained in a few-shot setting (200 samples). Results are included in Table~\ref{tab:analysis} as \textbf{D$^3$}*. It is slightly worse than our method with sufficient NLI training data, but still superior to most compared methods. Note that the response diversity metrics nearly remain unchanged. This means that our data diversification methods are still effective when starting from noisy distilled samples. It also shows that our method can be useful when only limited in-domain NLI labeled data are available for data distillation.
%The reason is that the distilled samples encourage the model to focus more on the personas while ignoring other aspects in dialogue. 
%Therefore, despite only using distilled data in training can promote \textbf{C} score, it significantly degenerates the model in other aspects. That is why we utilize curricula that cover the original data format.
% Several variants of our method is tested and their performance in terms of automatic metrics is shown in Table~\ref{tab:variants}. Here \textit{only augmented data} means the raw data $\mathcal{D}$ is not included in training but only $\mathcal{D}^a$ is used. \texit{w/o diversification} means the distilled data will directly used as $\mathcal{D}^a$ without introducing extra samples. \textit{w/o distillation} means same number of samples are randomly selected from raw data instead of distilled ones while other parts remains the same. It can be found that only using augmented data results in much lower n-gram accuracy while a very high \textbf{C} score as it tends to directly copy persona texts but ignore the coherence. Without using diversification results in a lower generation diversity as the distilled sample quantity is limited. In addition, without distillation affect the performance much as many noisy samples are included in training, resulting a higher uncertainty. The influence is neutralized by response filter which 
% \footnote{***not yet edited this para}

\begin{table*}[t!]
\setlength{\tabcolsep}{1.4mm}
\setlength{\abovecaptionskip}{0.2cm}
\setlength{\belowcaptionskip}{-0.2cm}
\centering
\small
\begin{tabular}{l|c|ccc|cccccc|c}
\toprule
 & PPL & BLEU & NIST-4 & BS$_f$ & Ent-1 & Ent-2 & Ent-3 & Dis-1 & Dis-2 & Dis-3 & \textbf{C} \\
\midrule
\textsc{Trans-\textbf{D$^3$}} & 37.30 & 3.358 & 1.206 & 0.1574 & 4.223 & 6.165 & 7.298 & 1.826 & 7.923 & 14.42 & 0.485 \\
\textit{Original} & 38.28 & 3.140 & 1.148 & 0.1486 & 4.046 & 5.484 & 6.262 & 1.609 & 6.298 & 11.71 & 0.235 \\
\textit{Only augment} & 126.3 & 1.603 & 0.956 & 0.0852 & 4.315 & 6.309 & 7.426 & 1.747 & 7.530 & 12.66 & 0.942\\
\textit{Shuffle} & 37.66 & 3.203 & 1.175 & 0.1521 & 4.128 & 6.096 & 6.979 & 1.659 & 6.889 & 13.79 & 0.404\\
\textit{Reverse} & 48.17 & 2.137 & 1.019 & 0.1508 & 3.947 & 5.291 & 6.039 & 1.368 & 5.503 & 9.211 & 0.912 \\
\bottomrule
\end{tabular}
\caption{\label{tab:curriculum_trans}Performance comparison between different curriculum variants, using \textsc{Transformer} as the base model.}
\end{table*}

\begin{table}[t!]
\setlength{\tabcolsep}{3mm}
\setlength{\abovecaptionskip}{0.2cm}
\setlength{\belowcaptionskip}{-0.2cm}
\small
\centering
\begin{tabular}{l|c|c|c|c}
\toprule
 & \multicolumn{4}{c}{Novelty-1, 2, 3, 4} \\
\midrule
sample & 30.89 & 47.07 & 53.81 & 59.64 \\
persona & 40.26 & 62.17 & 70.47 & 77.81 \\
% utterance & 26.20 & 39.52 & 45.48 & 50.56 \\
\bottomrule
\end{tabular}
\caption{\label{tab:novelty} Novelty metrics of the diversified data compared to distilled data in sample and persona level.}
\end{table}

\begin{figure}[!t]
\setlength{\abovecaptionskip}{0.2cm}
\setlength{\belowcaptionskip}{-0.2cm}
    \centering
    \includegraphics[width=0.65\linewidth]{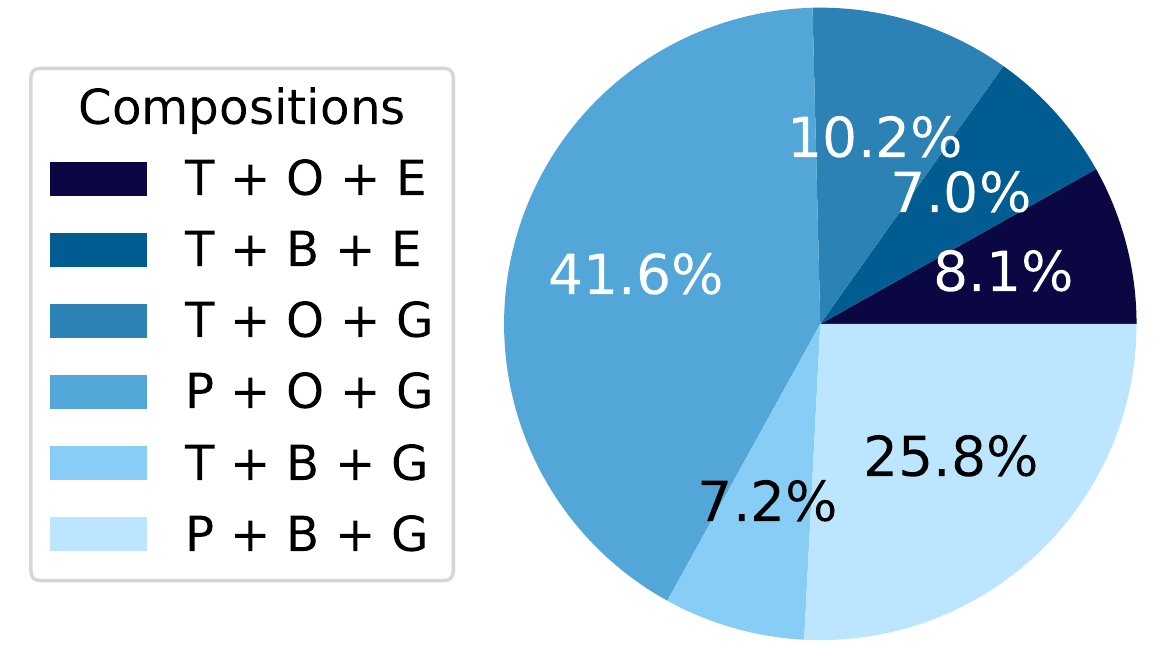}
    \caption{\label{fig:diversified_cases}The compositions of diversified data. T/P: token/phrase-level editing to get edited personas, O/B: original/BT-augmented dialogue history, E/G: token editing/generating by a model to get aligned responses.}
\end{figure}

\paragraph{Analysis of data diversification}
Table~\ref{tab:statistics} shows that the diversified data contain many new persona sentences as well as tokens. Besides, we compute the Novelty metrics~\cite{novelty_metrics,dialogue_distillation} of diversified samples in $\mathcal{D}^{div}$. It takes the original distilled samples in $\mathcal{D}^{dis}$ as references, and uses the Jaccard similarity function to measure the proportion of n-grams ($n=1,2,3,4$) in $\mathcal{D}^{div}$ but not in $\mathcal{D}^{dis}$. A higher value means more ``novel'' content. Note that we particularly prefer more novel personas, while not encouraging more novel dialogue histories. Thus, the Novelty scores on the overall samples which include dialogue histories, personas and responses, are lower than those on the personas.
%Results in Table~\ref{tab:novelty} again conclude the same observation.
% \footnote{***a reviewer asks u to say more about the novelty metric.}
% I have added

To further examine how each part of data diversification works, we conduct the following ablation studies: 1) \textit{w/o persona editing}: no persona sentence will be edited; 
2) \textit{w/o history augmentation}: only original dialogue history is used; 
3) \textit{w/o response filtering}: all constructed samples are directly used without using Eq.~\ref{eq:filter}.
Results in the bottom of Table~\ref{tab:analysis} show that all these designs contribute to the performance of the whole method. 
Among them, response filtering is the most important as it ensures the quality of augmented samples.
%Introducing new personas and paraphrased history are both beneficial for generation diversity. The former one has a significant effect on \textbf{C} score as novel persona texts benefit model robustness on persona consistency.
% \footnote{***not yet edited this para}

We also investigate the proportions of diversified samples coming from various source combinations.
Results are shown in Figure~\ref{fig:diversified_cases}, which shows that more than 80\% diversified samples have their responses obtained via model predicting, as token editing sets a strict condition that overlapped tokens must exist. 
Phrase-level editing also contributes to more high-quality personas with satisfactory fluency and semantic novelty.  

\paragraph{Analysis of data curriculum}
%To demonstrate the effectiveness of training with the designed data curriculum,
We first compare other data curriculum variants to show the usefulness of training with the designed data curriculum. The following variants are included: 1) \textit{Original}: only the original dataset $\mathcal{D}$ (the hard curriculum in \textbf{D$^3$}) is used, which is equal to the base model; 2) \textit{Only augment}: only the augmented dataset $\mathcal{D}^a$ (the easy curriculum in \textbf{D$^3$}) is used; 3) \textit{Shuffle}: shuffling of the original dataset $\mathcal{D}$ and the augmented dataset $\mathcal{D}^a$ together to train the model; 4) \textit{Reverse}: using the curricula in a reverse order, which means the hard curriculum first and then the easy one. 

Relevant results are shown in Table~\ref{tab:curriculum_trans}.
There is no doubt that our curriculum is the best when comprehensively considering all aspects. Although \textit{Only augment} and \textit{Reverse} show high C-scores, their responses are much worse in n-gram accuracy as they involve more persona information while focusing less on the dialogue coherence during generating. \textit{Shuffle} shows better performance than \textit{Original} as it includes more augmented data than the original dataset, which may benefit the training. However, such a mixing strategy is not so efficient as our data curriculum as it neglects the learning difficulty of different data sources.
%by 1) shuffling two kinds of data together (original data $\mathcal{D}$ and augmented data $\mathcal{D}^a$), and 2) using a reverse curriculum order.
%Our data curriculum obtains consistently the best performance among them on all metrics.
% Details can be found in Appendix~\ref{app:cirriculumn_analysis}.
%along with only using the original training data only (No Augmentation) or our curriculum strategy (Ours). 
%Obviously, Curr. is the best among four methods. Mix method has an intermediate performance between Raw and Curr. because the extra data benefits the model in data level but mixing reduce the learning efficiency as two kinds of data are not parallel. Rev. delivers results similar to training using only augmented data as the model tends to fit the distribution of later curriculum.
%

\begin{figure}[!t]
\setlength{\abovecaptionskip}{0.2cm}
\setlength{\belowcaptionskip}{-0.2cm}
    \centering
    \includegraphics[width=1\linewidth]{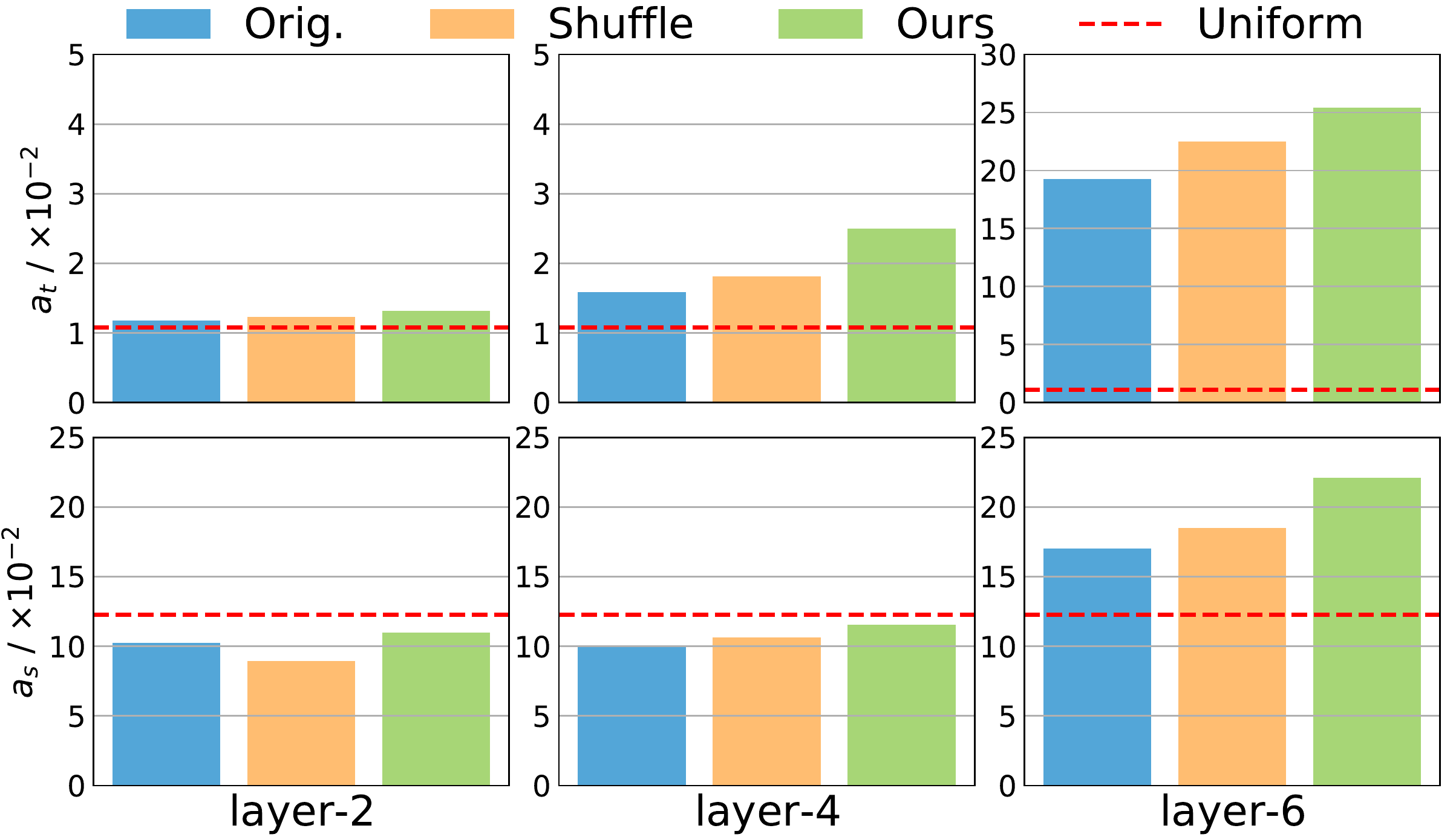}
    \caption{\label{fig:attention}Average consistent attention weights in different decoder layers of \textsc{Transformer} trained with (i) original dataset (Orig.), (ii) shuffled data in $\mathcal{D}$ and $\mathcal{D}^a$ (Shuffle), and (3) our data curriculum. Uniform: uniform attention values on all positions. Top: token-level $a_{t}$; bottom: sentence-level $a_{s}$. }
\end{figure}

Next, we further quantify the effect of curriculum training on models using the attention from the response on the persona sentences. 
We define two metrics, token-level/sentence-level consistent attention weight ($a_{t}$ and $a_{s}$), to measure how the attention contributes to reflecting the proper personas. 
Recall that we concatenate the persona sentences and history utterances as a single model input.
We record the token positions of the entailed persona sentences in the input sequence, which are determined by our NLI model, denoted as $\mathcal{S}$.
Then for each index $s \in \mathcal{S}$, if its corresponding token in the input also occurs in the response, we put this index pair into a set $\mathcal{T}=\{(s, l)\}$, where $s$ and $l$ are the token positions in the input sequence and response sequence respectively. Then we have two measurements for each sample:
% We first distill the personas of each sample.
% If the {k}-th persona text $P_k$ is predicted to entail the response (if exists), we record all indices of its tokens in the input sequence. A union of such positions in a sample is denoted as $\mathcal{S}$ for $a_{t}$. 
% % Then a union of such positions is denoted as $\mathcal{T}$.
% Then a set of matched token index pairs $\mathcal{T}=\{(x_l, y_l)\}$ is obtained, where $x_l \in \mathcal{S}$ and $y_l$ are index of input sequence and response respectively, and the token corresponds to $x_l$ matches the token in $y_l$.
\begin{equation}
    a_{t} = {\frac{1}{|\mathcal{T}|}\sum\limits_{(i,j) \in \mathcal{T}} {a_{ij}} }, \quad
    a_{s} = \frac{1}{Y}\sum\limits_{i = 1}^Y {\sum\limits_{j \in \mathcal{S}} {{a_{ij}}} },
\end{equation}
where $a_{ij} \in [ {0,1} ]$ is the normalized scalar attention weight at the $i$-th decoding step on the $j$-th input token, i.e. $\sum\nolimits_j {a_{ij}}=1$, and $Y$ is the length of the generated response.
A higher $a_{t}$/$a_{s}$ indicates that the model poses more attention on proper persona tokens, where the former one is fine-grained for reflecting how the attention works properly at each step, while the latter one is coarse-grained for the whole generated response.

Part of the results with selected \textsc{Transformer} layers for these two metrics on all samples from the PersonaChat dev set are shown in Figure~\ref{fig:attention} (Refer to Appendix~\ref{app:cirriculumn_analysis} for the complete results). Obviously, our method shows the highest $a_{t}$ and $a_{s}$ on all given layers compared to other two curriculum variants. Such a superiority is more significant in higher layers, which is more decisive for generating responses~\cite{fan2019reducing}. While the attentions weights tend to distribute uniformly in lower layers, which are close to the uniform values.

\begin{figure}[!t]
\setlength{\abovecaptionskip}{0.2cm}
\setlength{\belowcaptionskip}{-0.2cm}
    \centering
    \includegraphics[width=1\linewidth]{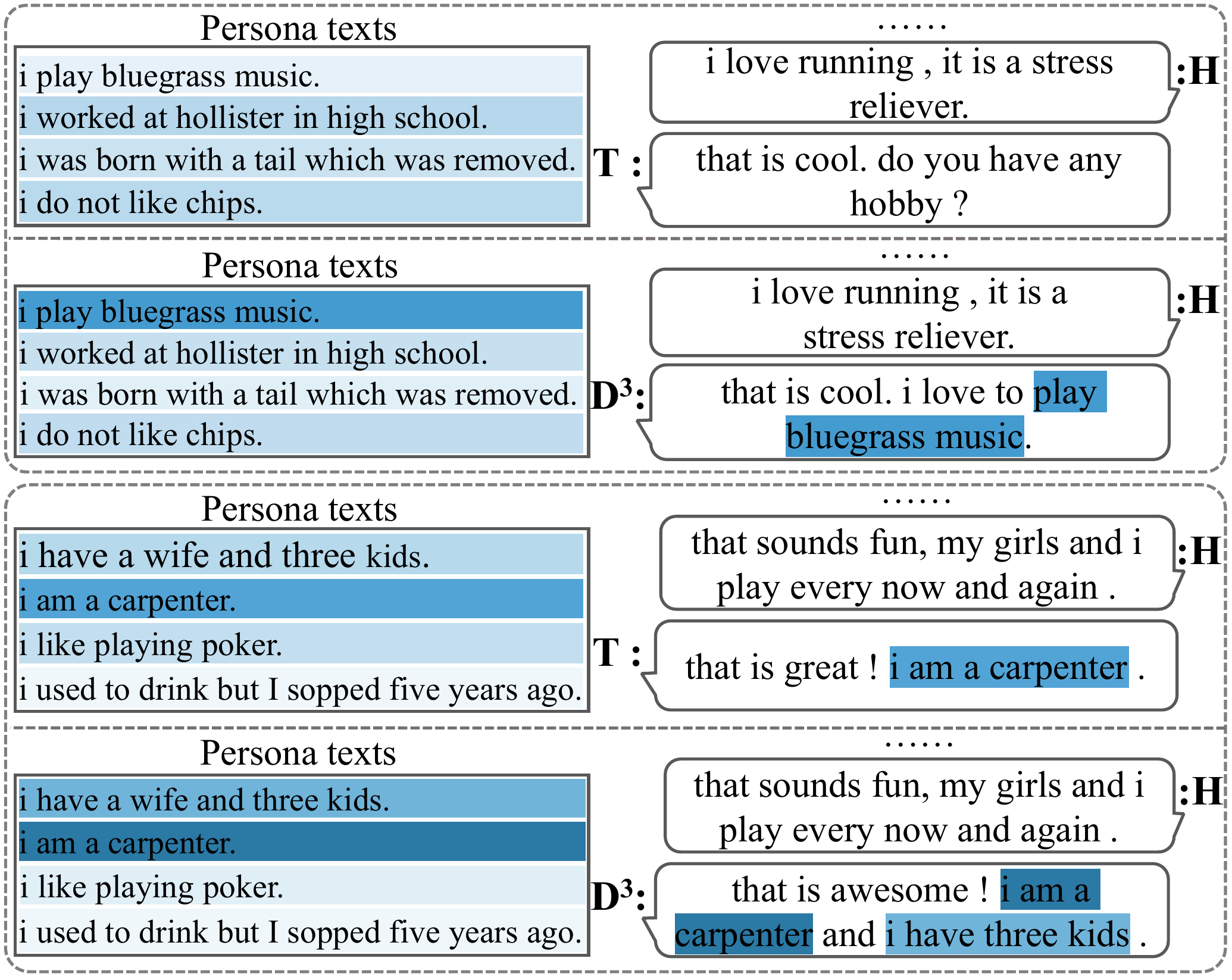}
    \caption{\label{fig:case}Sample responses and visualized model attention weights on personas texts ($a_s$), deeper colors indicate higher attention weights. \textbf{T}:\textsc{Transformer}, \textbf{D$^3$}:\textsc{Transformer-\textbf{D$^3$}}. }
\end{figure}

\paragraph{Case study} Some response samples generated when using \textsc{Transformer} as the base model are shown in Figure~\ref{fig:case}. Here \textbf{H} indicates dialogue history, a persona sentence shaded in a darker color denotes that it has a higher attention weight posed by the model. Our method \textbf{D$^3$} can offer a model with the capability to pose more attention on the proper persona texts during generating responses. More cases can be found in Appendix~\ref{app:more_cases}.
% We also visualize the attention weights on different personas during generating responses. 
% The darker color on persona texts denotes a higher attention weight. 

\section{Conclusion}

Our work targets the challenging personal-based dialogue generation task. Unlike previous work that designs a new dialogue model to improve the generation performance, we analyze the data issues affecting current models. On one hand, the data scale and diversity are expensive to increase by data collection. On the other hand, current data are difficult to learn with. Based on such an understanding, we propose a model-agnostic data manipulation method for this task. It first distills the original data and then augments both the amount and diversity of the distilled data. A curriculum training is then applied to utilize both augmented and original data. 
%It aims for more effectively model attention learning on proper persona profile among multiple candidates so as to generate better responses. Raw data is distilled to get samples with determinate consistency between a persona text and the response. In order to remedy the limited quantity and diversity, they are diversified via multiple model-based approaches, along with filtering covering several aspects to ensure quality. 
% Finally,  we train a model by a curriculum strategy that the distilled data including both the original distilled and the augmented ones are learned first, and then followed by the original training samples. 
Experimental results showed that our method effectively improves the performance of two strong dialogue models, i.e. Transformer encoder-decoder and GPT2.

\section*{Acknowledgements}
We would like to thank Piji Li and Lemao Liu for their helpful discussion and feedback. We also thank anonymous reviewers for their constructive comments.

\bibliography{anthology,custom}
\bibliographystyle{acl_natbib}

\newpage
\appendix

\section{Implementation Details of \textbf{D$^3$}}

\subsection{\label{app:detail_distillation}Details of Distillation}

In order to obtain the NLI model to determine the persona consistency, the RoBERTa-Large-MNLI\footnote{https://huggingface.co/roberta-large-mnli} model is utilized. To make the model better fit the domain of PersonaChat, we finetune the model on the DialogueNLI dataset~\cite{dialogue_nli} which is a part of the original PersonaChat.
We set the batch size as 32 and finetune the model for 5 epochs using a learning rate 1e-5. 
We obtain a model RoBERTa$_{nli}$ achieving 90.8\% accuracy on the dev set. 
This model will also be responsible for calculating the entailment probability $\text{NLI}$ in response filtering and \textbf{C}-score in the experiments. 
A threshold $\tau=0.99$ is used in this model for predicting the NLI labels. For the few-shot setting \textbf{D$^3$}* in \secref{sec:more_analysis}, we randomly sample 200 samples from the training set to train the above NLI model using learning a rate 2e-5, and obtain a model achieving 79.3\% on the dev set.
% In addition, to keep the confidence of distilled samples, we set a threshold $\tau=0.99$ for entailment probability $p_e$ given by the NLI model. 

\subsection{\label{app:data_diversification}Details of Diversification}

The BERT-based-uncased model\footnote{https://huggingface.co/bert-base-uncased} and GPT2-base\footnote{https://huggingface.co/gpt2} are involved as the pretrained models in this stage. To ensure that the pretrained models can make predictions that better fit current data domain while also have enough capabilities of generation diversity, we perform the following finetuning: 1) finetune BERT and GPT2 on the persona sentences for 100 steps with a batch size 32 and a learning rate 1e-4, obtaining BERT$_{per}$ and GPT2$_{per}$; 2) finetune GPT2 on responses for 200 steps with a batch size 32 and a learning rate 1e-4, and obtain GPT2$_{res}$.

\paragraph{Persona editing} BERT$_{per}$ and GPT2$_{per}$ will be used for token-level editing and phrase-level editing respectively. Each will generate 10 unique new persona sentences from one original persona sentence via sampling according to the multinomial distribution. At the token level, we only mask the most informative tokens which can be decided by the POS tags given by SpaCy\footnote{https://spacy.io/} as it is meaningless to mask some words such as prepositions ``to" and ``in". The target POS tags are listed in Table~\ref{tab:target_pos}. We set the token-level mask ratio as 0.8. At phrase level, the mask ratio is randomly sampled between $[0.3, 0.6]$. We also restrict that at least 2 tokens are masked and the maximum length of generated text pieces from GPT2$_{per}$ does not exceed 30\% of the original length to preserve the sentence similarity.

\begin{table}[t!]
\setlength{\abovecaptionskip}{0.2cm}
\setlength{\belowcaptionskip}{-0.2cm}
\centering
 \small
\begin{tabular}{l|c}
\toprule
% ~ & Token-level masking \\
% \midrule
POS tags & {\tabincell{c}{VERB, NOUN, PROPN, NUM,\\ ADV, ADP, ADJ}} \\
\bottomrule
\end{tabular}
\caption{\label{tab:target_pos} The target POS tags for token-level masking.}
\end{table}

\begin{table*}[t!]
\setlength{\abovecaptionskip}{0.2cm}
\setlength{\belowcaptionskip}{-0.2cm}
\centering
\small
\setlength{\tabcolsep}{1.1mm}
\begin{tabular}{l|c|ccc|cccccc|c}
\toprule
~ & PPL & BLEU & NIST-4 & BS$_f$ & Ent-1 & Ent-2 & Ent-3 & Dis-1 & Dis-2 & Dis-3 & \textbf{C} \\
\midrule
Multi-GPT2 & 17.70 & 6.186 & 1.4773 & 0.3216 & 4.665 & 6.809 & 7.704 & 4.111 & 15.693 & 27.115 & 0.850 \\
\bottomrule
\end{tabular}
\caption{\label{tab:multi_gpt2} The performance of trained Multi-GPT2 on the distilled dev set. }
\end{table*}

We use $\alpha=0.4$ in Eq.~1, where $\text{PPL}$ is given by GPT2$_{per}$ normalized by a constant $50$ (which is about the highest PPL value given by the GPT2 model on current corpus). For BERTScore, the F1 value is used as $BS_{f}$ while other configurations follow the recommendation for English in \citet{bert_score}\footnote{https://github.com/Tiiiger/bert\_score}. $N_p$ is set as 5.
%which means 5 new personas with the lowest $s_p$ originated from the same original persona are remained in ${\widetilde P}^d$. 
%Note that we obtain edited personas for each unique distilled persona sentence rather than each distilled sample.

\paragraph{Response aligning} 
%Given the permutations of pseudo personas and dialog history utterances from different sources, we only apply token-level editing on persona-history pairs whose source distilled sample contains consistent tokens exist between ${\widetilde P}$ and ${\widetilde H}$. 
For token-level editing,
we also restrict the POS tags of overlapped tokens according to Table~\ref{tab:target_pos}. 
%Then editing will be processed on the corresponding positions in the original responses, replacing old tokens with new ones to get aligned responses. 
For model predicting, we train the Multi-GPT2 model on the distilled data $\mathcal{D}^{dis}$. Its performance on the dev set distilled from the original dev set of PersonaChat is shown in Table~\ref{tab:multi_gpt2}. We can see that this model shows high n-gram accuracy and persona consistency, thus should be effective. 

\paragraph{Dialogue history augmentation} We use the \textit{transformer\_wmt\_en\_de} Transformer model in Fairseq\footnote{https://github.com/pytorch/fairseq} as the translation model. It is trained on the WMT14 EN-FR dataset with 40.5M samples and default configurations.
%All configurations follow the default ones and the training step number is 10000. 
During inference, we use beam search with its size 5 for both en-fr and fr-en translation, resulting in 25 new utterances for each original one. For a large divergence, we select $N_p=1$ new utterance with the lowest BLEU score when taking the original one as the reference.

\paragraph{Quality filtering} 
We use GPT2$_{res}$ normalized by a constant $50$ to get the PPL of responses. 
%Based on the previous study that a NLI model can also be used to determine the coherence between utterances~\cite{nli_coherence}
Here, we finetune another RoBERTa-Large-MNLI model on the InferConvAI dataset\footnote{https://github.com/nouhadziri/DialogEntailment} which achieves 88.7\% accuracy on its dev set. The entailment probability given by this model is regarded as $\text{NLI}_{c}$. We set $\beta = 0.2$, $\gamma = 0.6$ in Eq.~2.

%\noindent \textbf{Quality of diversified samples} 
%To prove the quality of generated responses in diversification, we employ (similar as Filtering) to measure its fluency and coherence to query respectively. 
We compare the fluency and coherence of responses with the GPT2-based PPL and NLI model-based score from the training set, which are shown in Table~\ref{tab:response_quality}. 
In addition, we also evaluate the GPT2-PPL's for edited and original persona sentences, which are 6.427 vs. 10.426.

\begin{table}[t!]
\setlength{\abovecaptionskip}{0.2cm}
\setlength{\belowcaptionskip}{-0.2cm}
\centering
\small
\setlength{\tabcolsep}{1.1mm}
\begin{tabular}{l|c|c}
\toprule
~ & GPT2-PPL & Coherence score \\
\midrule
\textbf{Original} & 13.119 & 0.361 \\
\textbf{Diversified} & 18.847 & 0.525 \\
\bottomrule
\end{tabular}
\caption{\label{tab:response_quality} The average GPT2-based PPL and NLI model-based coherence score of the original responses and responses generated in diversification. }
\end{table}

\section{\label{app:details_experiments}Details of Experiment}

\paragraph{Base model} For \textsc{Transformer}, we use 300-dim GloVe~\cite{glove} trained on 6B corpus as the word embeddings. There are 6 layers in both the encoder and decoder, with the hidden size 300 and 4 heads. During training, a cross-entropy loss is used along with Label Smoothing with the ratio 0.1. 
For GPT2, we use the base pretrained model with 12 layers and 768-dim hidden state. It will be trained using the average of a cross-entropy loss on generating and a classification loss between true response and one randomly sampled negative response. 
Beam search with the beam size 3 along with length penalty is used during inference for both models. 

\begin{figure*}[ht]
\setlength{\abovecaptionskip}{0.2cm}
\setlength{\belowcaptionskip}{-0.2cm}
    \centering
    \includegraphics[width=0.8\textwidth]{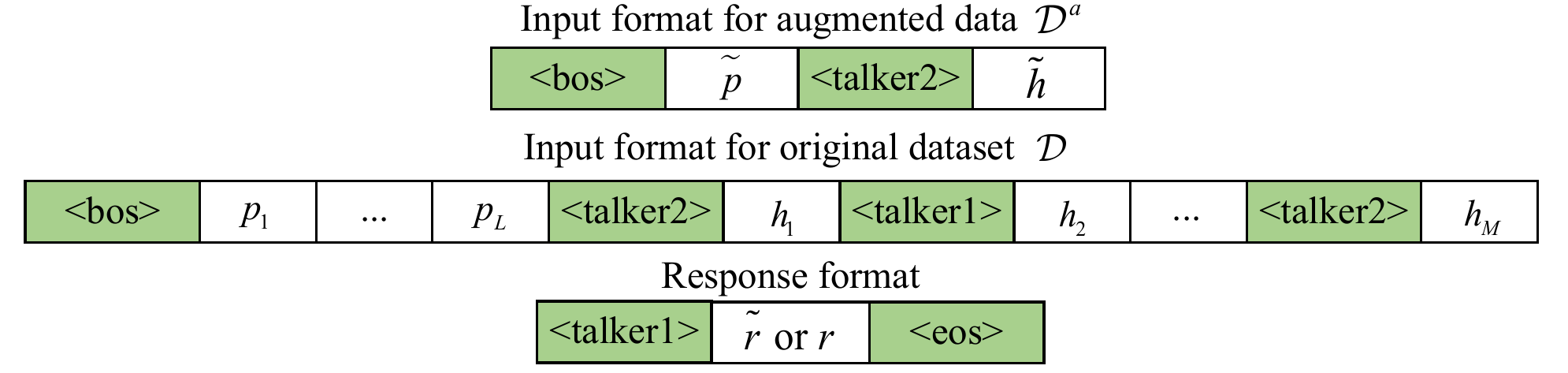}
    \caption{\label{fig:data_format}The sequence format of an input and an output for both \textsc{Transformer} and GPT2 models. }
\end{figure*}

The formats of input or response for both models are shown in Figure~\ref{fig:data_format}. 
Here $<$bos$>$, $<$eos$>$, $<$talker1$>$, and $<$talker2$>$ are special symbols to distinguish different parts of input or response. 
%For an augmented sample $(P^a, H^a, R^a)$, $P^a$, $H^a$ and $R^a$ only contain a single persona text $p_a$, a single history utterance $h_a$ and a single response $r_a$ respectively.

\paragraph{Model training} We use a learning rate 2e-4 for \textsc{Transformer} and 6.25e-5 for GPT2, which is a common setting in former similar works. And the training batch size is 256 for both models. Training will be stopped until the loss on the dev set does not decrease for $N$ epochs. Here $N$ is 15 for \textsc{Transformer} and 5 for GPT2. In curriculum learning, the learning rate is the same for different curricula. The dev set of the easy curriculum is obtained by applying the same augmentation to the original dev set.  Models with the minimum loss at each curriculum are remained as the best. The best model obtained on the easy curriculum is used as the initial model in the hard curriculum. All experiments are implemented via PyTorch on 32GB NVIDIA V100 GPUs. Each epoch takes about 10 min for Transformer and 25min for GPT2. 

\paragraph{Hyper-parameters} All hyper-parameters are determined using a coarse grid search to ensure satisfactory performance, including $\tau$ in data distillation, $\alpha$ in Eq.~1, $\beta, \gamma$ in Eq.~2. The candidate values of these hyper-parameters are given in Table~\ref{tab:hyper_parameter}, which are determined empirically to reduce the searching cost. The search target we want to maximize is the normalized average of all automatic metrics listed in Table~\ref{tab:main_result} when inferencing on the test set, except PPL. Note that we only take \textsc{Transformer} as the base model for search, each time of search takes about 0.7 GPU day. GPT2 model follows the same setting as \textsc{Transformer}. We found that $\tau$ plays a more important role in our method who determine the quality of distilled samples, while other parameters have fewer impacts on our method.

\begin{table}[t!]
\setlength{\abovecaptionskip}{0.2cm}
\setlength{\belowcaptionskip}{-0.2cm}
\centering
\small
\setlength{\tabcolsep}{1.1mm}
\begin{tabular}{l|c}
\toprule
Param & Candidate values \ \\
\midrule
$\tau$ & 0.9, 0.95, 0.99 \\
$\alpha$ & 0.4, 0.5, 0.6 \\
$\beta$ & 0.2, 0.3 \\
$\gamma$ & 0.4, 0.5, 0.6 \\
\bottomrule
\end{tabular}
\caption{\label{tab:hyper_parameter} The candidate values for hyper-parameters during grid searching.}
\end{table}

\begin{table}[t!]
\setlength{\abovecaptionskip}{0.2cm}
\setlength{\belowcaptionskip}{-0.2cm}
\centering
\small
\setlength{\tabcolsep}{1.1mm}
\begin{tabular}{l|r}
\toprule
method & Train sample number \ \\
\midrule
Original & 65,719 \\
\textbf{BT} & 131,436 \\
\textbf{CVAE} & 131,436 \\
\textbf{Entropy-Filter} & 59,892 \\
\textbf{\textbf{D$^3$}}(Ours) & {\tabincell{c}{53,393 (easy) \\ 65,719 (hard) \\ 119,112 (all)}}\\
\bottomrule
\end{tabular}
\caption{\label{tab:data_num} The training sample number used in each method.}
\end{table}

\paragraph{Baselines} We apply the same translation models as the ones used in \secref{app:data_diversification} for the BT~\cite{back_translation} baseline and augment each sample with a new sample from it.
%, extending each sample with a new sample originated from it which is consisted of texts that have the lowest BLEU scores to the original one.
For CVAE~\cite{augmentation_cvae} method, we use its default setting to train the model on PersonaChat dataset without using the personas. A new sample is generated for each input in the original dataset. In Entropy-filter~\cite{augmentation_filter}, we set the threshold as 1.1 and using both source and target sequences for filtering. Only samples that survived after filtering are used in training. The total numbers of training samples of all methods are listed in Table~\ref{tab:data_num}. Note that 0all models are trained until the loss does not decrease for the same $N$ epochs for a fair comparison.

\paragraph{Metrics} 
We use the same $\text{BS}_f$ and RoBERTa$_{nli}$ obtained before to calculate the BERTScore and C-score metrics respectively. The instructions for human annotators are provided in Table~\ref{tab:instruction1} and \ref{tab:instruction2}.
%
%The BERTScore presented in our experiments is the F1 score implemented using the default setting and official script with rescaling\footnote{https://github.com/Tiiiger/bert\_score}. RoBERTa$_{nli}$ obtained before is used here to calculate \textbf{C score}.

\begin{table*}[t!]
\setlength{\tabcolsep}{1.4mm}
\setlength{\abovecaptionskip}{0.2cm}
\setlength{\belowcaptionskip}{-0.2cm}
\small
\centering
\begin{tabular}{l|c|ccc|cccccc|c}
\toprule
 & PPL & BLEU & NIST-4 & BS$_f$ & Ent-1 & Ent-2 & Ent-3 & Dis-1 & Dis-2 & Dis-3 & \textbf{C} \\
\midrule
\textsc{GPT2} & 17.63 & 3.761 & 1.278 & 0.1693 & 4.485 & 6.187 & 7.029 & 2.011 & 8.260 & 15.03 & 0.518  \\
\textsc{GPT2-\textbf{D$^3$}} & 15.69 & 4.184 & 1.429 & 0.1835 & 4.614 & 6.426 & 7.321 & 2.179 & 9.458 & 17.72 & 0.557 \\
\textsc{GPT2-\textbf{D$^3$}*} & 15.77 & 4.082 & 1.388 & 0.1809 & 4.611 & 6.408 & 7.312 & 2.209 & 9.657 & 17.91 & 0.536 \\
\midrule
\quad \textit{w/o diversification} & 15.89 & 4.119 & 1.441 & 0.1817 & 4.526 & 6.281 & 7.148 & 2.131 & 9.243 & 17.11 & 0.528 \\
\quad\quad \textit{w/o distilled format} & 16.04 & 4.026 & 1.379 & 0.1788 & 4.462 & 6.151 & 7.097 & 2.017 & 9.022 & 16.86 & 0.518 \\
\quad\quad \textit{only distillation} & 29.73 & 2.912 & 1.325 & 0.1509 & 4.558 & 6.392 & 7.250 & 1.252 & 4.807 & 9.048 & 1.131 \\
\midrule
\quad \textit{w/o persona editing} & 15.81 & \textit{4.190} & 1.427 & 0.1801 & 4.503 & 6.204 & 7.062 & 2.065 & 8.867 & 16.83 & 0.524 \\
\quad \textit{w/o history augmentation} & 15.75 & \textit{4.213} & \textit{1.503} & 0.1812 & 4.562 & 6.333 & 7.244 & 2.057 & 9.131 & 17.34 & 0.533 \\
\quad \textit{w/o response filte}r & 15.83 & 4.119 & 1.395 & 0.1790 & 4.604 & 6.387 & 7.265 & 2.158 & 9.414 & \textit{17.74} & 0.518 \\
\bottomrule
\end{tabular}
\caption{\label{tab:analysis_distill}Automatic evaluation results with variant settings in  distillation variants (middle), and data diversification ablations (lower), compared with the original \textbf{D$^3$}(top) on GPT2. \textbf{D$^3$}* means using an NLI model trained under a few-show setting (200 labelled samples) in the data distillation.}
\end{table*}

\begin{table*}[t!]
\setlength{\tabcolsep}{1.4mm}
\setlength{\abovecaptionskip}{0.2cm}
\setlength{\belowcaptionskip}{-0.2cm}
\centering
\small
\begin{tabular}{l|c|ccc|cccccc|c}
\toprule
& PPL & BLEU & NIST-4 & BS$_f$ & Ent-1 & Ent-2 & Ent-3 & Dis-1 & Dis-2 & Dis-3 & \textbf{C} \\
\midrule
\textsc{GPT2-\textbf{D$^3$}} & 15.69 & 4.184 & 1.429 & 0.1835 & 4.614 & 6.426 & 7.321 & 2.179 & 9.458 & 17.72 & 0.557 \\
\textit{Orignal} & 17.63 & 3.761 & 1.278 & 0.1693 & 4.485 & 6.187 & 7.029 & 2.011 & 8.260 & 15.03 & 0.518 \\
\textit{Only augment} & 33.01 & 2.540 & 1.078 & 0.1035 & 4.574 & 6.255 & 7.232 & 1.916 & 7.340 & 11.77 & 1.148 \\
\textit{Shuffle} & 16.58 & 3.801 & 1.321 & 0.1799 & 4.588 & 6.261 & 7.216 & 2.128 & 9.391 & 17.55 & 0.525 \\
\textit{Reverse} & 30.46 & 2.615 & 1.069 & 0.1189 & 4.298 & 6.074 & 6.960 & 1.646 & 6.709 & 9.529 & 1.111 \\
% No Aug. & 17.63 & 3.761 & .8598 & 0.518 & 5.904 & 8.435 \\
% Aug. & 33.01 & 2.540 & .8487 & 1.148 & 6.035 & 4.949 \\
% Shuffle & 16.58 & 3.801 & .8616 & 0.525 & 6.032 & 9.632 \\
% Reverse. & 30.46 & 2.615 & .8513 & 1.111 & 5.787 & 4.097 \\
% Ours & \bf 15.69 & \bf 4.184 & \bf .8622 & \bf 0.557 & \bf 6.122 & \bf 9.787 \\
\bottomrule
\end{tabular}
\caption{\label{tab:curriculum_gpt2}Performance comparison between different curriculum variants, using GPT2 as the base model.}
\end{table*}

\section{Additional Experimental Results}

\subsection{\label{app:attention_history}Attention on Dialogue History}

\begin{figure}[t!]
\setlength{\belowcaptionskip}{-0cm}
    \centering
    \includegraphics[width=1\linewidth]{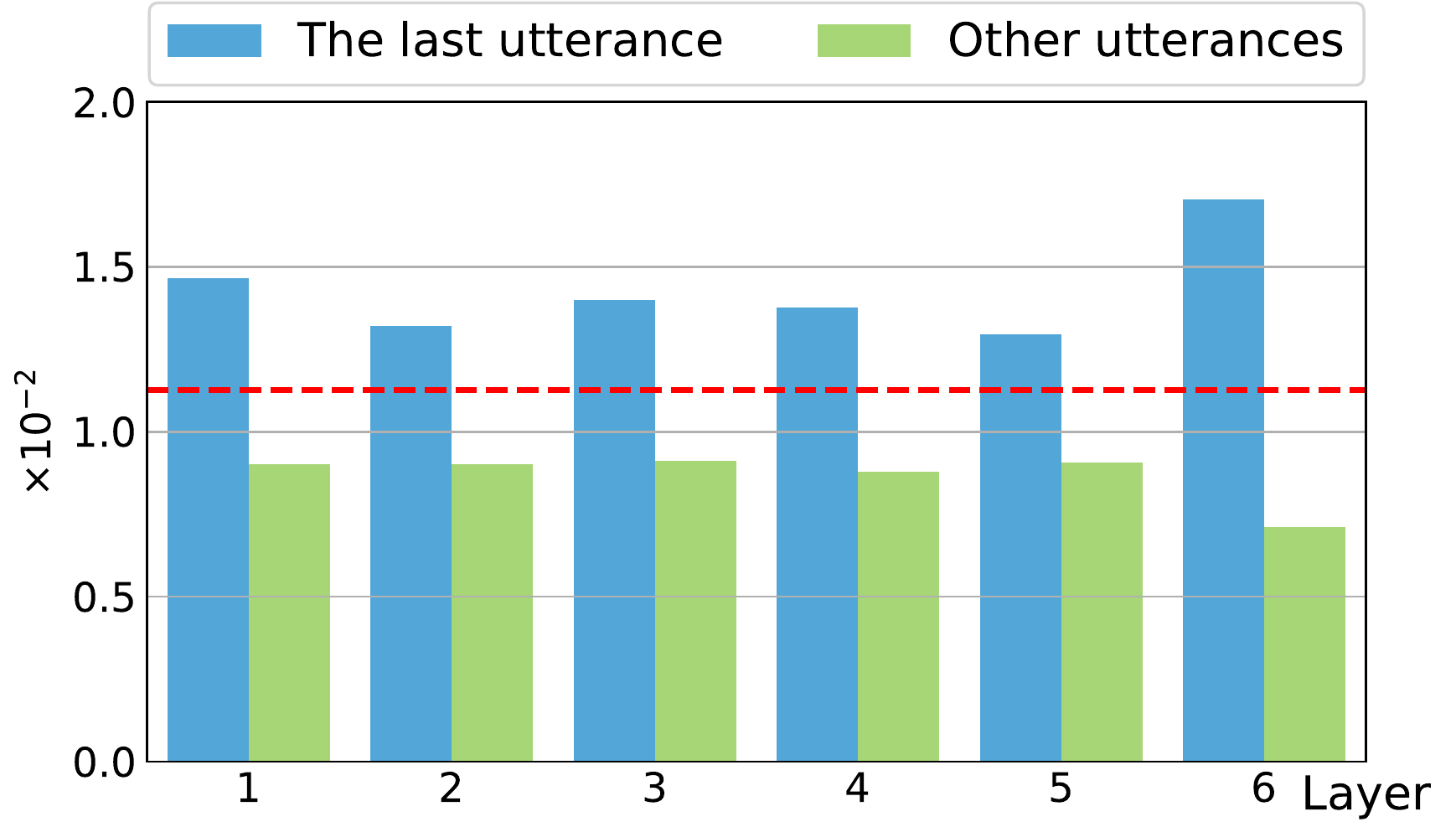}
    \caption{\label{fig:history_attention}The average token-level attention weights from different decoder layers in \textsc{Transformer} on the last utterance or other part of dialogue history. Red line: the baseline values when all attention distributes uniformly among all tokens.}
\end{figure}

To investigate how models pose attention on each part of dialogue history, especially the last utterance, we calculate the attention weights from different decoder layers on the last utterance or the other dialogue history utterances. \textsc{Transformer} model is used here, which is trained with the original training data without any augmentation. 
When testing on the dev set of PersonaChat dataset, the average token-level attention weight on the last utterance in the dialogue history is significantly higher than that on all other utterances, as shown in Figure~\ref{fig:history_attention}.
%Obviously, the last utterance obtains more attention, while other parts obtain less than the average value. 
%The reason is that the last utterance usually has a higher coherence with the response, the model also learns to fit this property to avoid noise. 
Thus, our history distillation can ease model learning for such knowledge by removing former utterances. 
% The sentence-level attention is the summation of all attention weights within this sentences, while the token-level value is the average of weights among all tokens. Results are shown in Figure~\ref{fig:history_attention}., obtained on the dev set of PersonaChat. 

% \begin{table*}[t!]
% \setlength{\tabcolsep}{1.2mm}
% \setlength{\abovecaptionskip}{0.2cm}
% \setlength{\belowcaptionskip}{-0.6cm}
% \small
% \centering
% \begin{tabular}{l|cccccc}
% \toprule
% \textbf{Model} & BS$_f$ & Dist-3 & \textbf{C} & Flu. & Coh. & Pcon. \\
% \midrule
% Trans & & & 1.4e$^-10$ & 0.0085 & 0.0236 & 2.8e$^-5$ \\
% Trans-BT & & & 7.79e$^-5$ & 0.0239 & 0.0689 & 0.0017 \\
% Trans-CVAE & & & 1.19e$^-5$ & 0.0253 & 0.0059 & 8.3e$^-5$ \\
% Trans-filter & & & 0.0282 & 0.0062 & 0.0392 & 0.0470 \\
% \midrule
% GPT2 & & & 0.0490 & 0.3597 & 0.2324 & 0.2564  \\
% GPT2-BT & & & 0.0091 & 0.0435 & 0.2492 & 0.0207 \\
% GPT2-CVAE & & & 0.0021 & 0.0002 & 0.0342 & 0.0016 \\
% GPT2-filter & & & 0.2549 & 0.2759 & 0.3720 & 0.2047 \\
% \bottomrule
% \end{tabular}
% \caption{\label{tab:p_value}The p values in T-test between our \textbf{D$^3$} and other baselines on two base models, in terms of different metrics.}
% \end{table*}

\subsection{\label{app:statistics_results}Statistical Results of Table~2}

We conduct Student's T-test between the experimental results of our method \textbf{D$^3$}) and every other baseline under each base model to verify the performance difference significance between every two methods. Here, all human evaluation results (Fluency, Coherence, Persona-consistency), and some applicable automatic metrics (C-score, BS$_f$) are included. 
% The details of p values are given in Table~\ref{tab:p_value}. 
We can find that nearly all results from baselines satisfy the null hypothesis (results are significantly different from \textbf{D$^3$}) given $p > 0.05$ or even a smaller theshold using \textsc{Transformer} as the base model. Such significant difference tends to appear fewer times when using GPT2 as the base model except for CVAE, which again shows that all data manipulation methods may have fewer impacts when packed with a pretrained model.

\subsection{\label{app:gpt2_ablation}More Analysis on GPT2}

We also provide the extensive analysis results on GPT2 which is similar to the ones given in \secref{sec:more_analysis} on \textsc{Transformer}. Table~\ref{tab:analysis_distill} shows the results. We can find the influence of data diversification, as well as our distillation, have fewer impacts on GPT2 compared to \textsc{Transformer}. The reason is that GPT2 is a strong pretrained model, being less vulnerable to the different numbers of data samples. Moreover, Table~\ref{tab:curriculum_gpt2} shows the performance when using different curriculum variants, demonstrating the similar conclusion as \textsc{Transformer.}

%The results of ablation studies in the data diversification module on GPT2 are shown in Table~\ref{tab:analysis_distill}. The performance gaps between them are also narrowed compared to the results when using Transformer as the base model. But the similar conclusions can still be drawn that response filter has a relatively more important contribution, while persona editing affects the generation diversity as well as persona consistency. History augmentation has the least significant influence.

\begin{figure*}[t!]
\setlength{\abovecaptionskip}{0.2cm}
\setlength{\belowcaptionskip}{-0.2cm}
    \centering
    \begin{subfigure}{0.95\linewidth}
        \centering
        \setlength{\belowcaptionskip}{0.2cm}
        \includegraphics[width=1\linewidth]{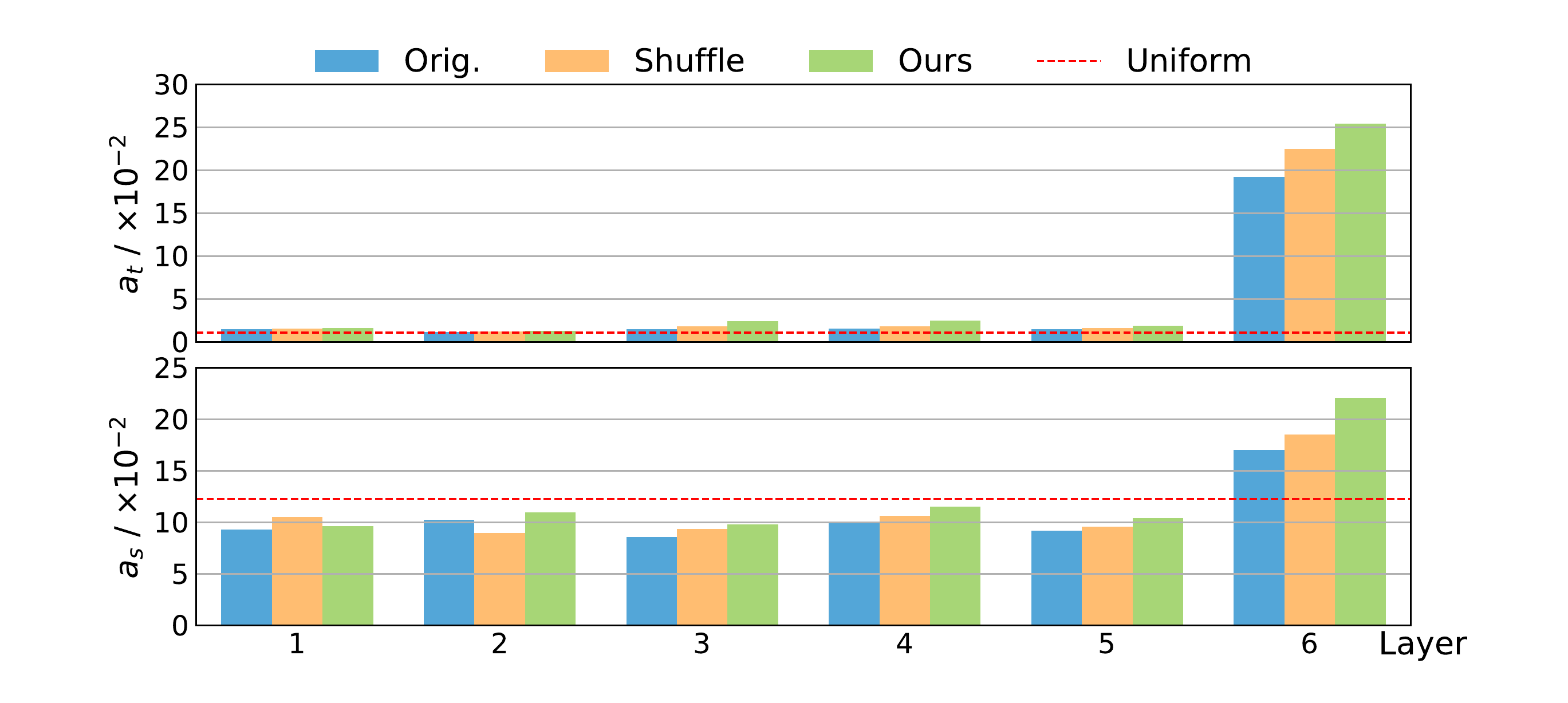}
        \caption{Consistent attention weights from different decoder layers in \textsc{Transformer}. Upper: token-level $a_{tc}$, lower: sentence-level $a_{sc}$.}
    \end{subfigure}
    \begin{subfigure}{0.95\linewidth}
        \centering
        \includegraphics[width=1\linewidth]{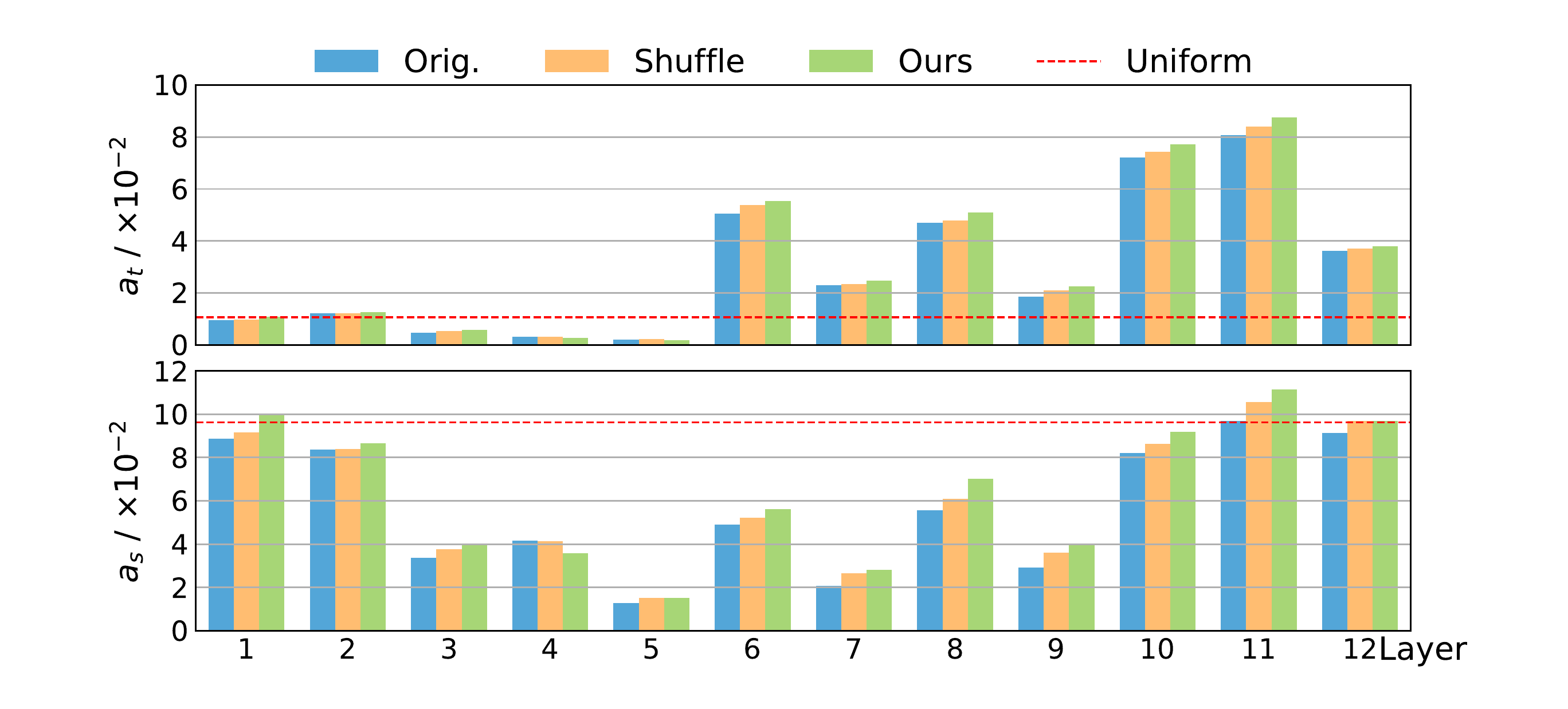}
        \caption{Consistent attention weights from different decoder layers in \textsc{GPT2}. Upper: token-level $a_{tc}$, lower: sentence-level $a_{sc}$.}
    \end{subfigure}
    \caption{\label{fig:attention_all}Consistent attention weights on \textsc{Transformer} and GPT2. Orig.:training the model using the original training data $\mathcal{D}$; Shuffle: training the model using the shuffling data of $\mathcal{D}$ and $\mathcal{D}^a$; Ours: training the model using our curriculum strategy; Uniform.: the attention value distributed on all positions uniformly, which is a baseline.}
\end{figure*}

\subsection{\label{app:cirriculumn_analysis}Additional Results of Attention Analysis for Curriculum}

\begin{table*}[t!]
\setlength{\tabcolsep}{1mm}
\setlength{\abovecaptionskip}{0.2cm}
\setlength{\belowcaptionskip}{-0.2cm}
\centering
\small
\begin{tabular}{l|c|ccc|cccccc|c}
\toprule
 & PPL & BLEU & NIST-4 & BS$_f$ & Ent-1 & Ent-2 & Ent-3 & Dis-1 & Dis-2 & Dis-3 & \textbf{C} \\
\midrule
\textsc{Trans-\textbf{D$^3$}} & 37.30 & 3.358 & 1.206 & 0.1574 & 4.223 & 6.165 & 7.298 & 1.826 & 7.923 & 14.42 & 0.485 \\
\textsc{Trans-\textbf{D$^3$}}(200\%) & 37.49 & 3.367 & 1.199 & 0.1570 & 4.271 & 6.235 & 7.343 & 1.821 & 7.997 & 14.51 & 0.493 \\
\textsc{Trans-\textbf{D$^3$}}(50\%) & 37.75 & 3.269 & 1.167 & 0.1551 & 4.132 & 6.085 & 7.003 & 1.743 & 7.658 & 14.10 & 0.468 \\
\bottomrule
\end{tabular}
\caption{\label{tab:div_number}Performance comparison between original \textbf{D$^3$} and variants when using diversified dataset $\mathcal{D}^{div}$ with about 200\% or 50\% size of distilled dataset $\mathcal{D}^{dis}$. }
\end{table*}

To better illustrate the effect of our training curriculum strategy, we further provide the token-level/ sentence-level consistent attention weights $a_t$ and $a_s$ in all layers of Transformer and GPT2 trained via 3 curriculum strategies, \textit{Original} (Orig.)., \textit{Shuffle} or our \textbf{D$^3$} method, as described in \secref{sec:more_analysis}.
All visualized attention weights are shown in Figure~\ref{fig:attention_all}. Our method has the most accurate attention on personas at both levels. On the other hand, compared to Transformer, the divergence between different layers in GPT2 is more significant.

\subsection{The Influence of Diversified Sample Numbers}

Since we can simply control the threshold for $s$ in Eq.~\ref{eq:filter} to determine how many diversified samples are generated for $\mathcal{D}^{div}$. How this quantity affect the performance of \textbf{D$^3$} ? We carry out experiments to use different $\mathcal{D}^{div}$ whose size is about 50\% of $\mathcal{D}^{dis}$ or 200\% of $\mathcal{D}^{dis}$ on \textsc{Transformer}, compared to the original method where $\mathcal{D}^{div}$ is nearly the same size as $\mathcal{D}^{dis}$. The results in terms of automatic metrics are shown in Table~\ref{tab:div_number}. It can be found that further extending the data scale will result in a very slight promotion but a longer training time, while squeeze the diversified dataset size has a more obvious effect on the performance. Nevertheless, using $\mathcal{D}^{div}$ with a similar size as $\mathcal{D}^{dis}$ is a good trade-off between resource cost and performance, while ensure a fair comparison between former methods.

\subsection{\label{app:more_cases}Additional Case Studies}

\begin{figure*}[t!]
    \centering
    \begin{subfigure}{0.49\linewidth}
        \centering
        \includegraphics[width=1\linewidth]{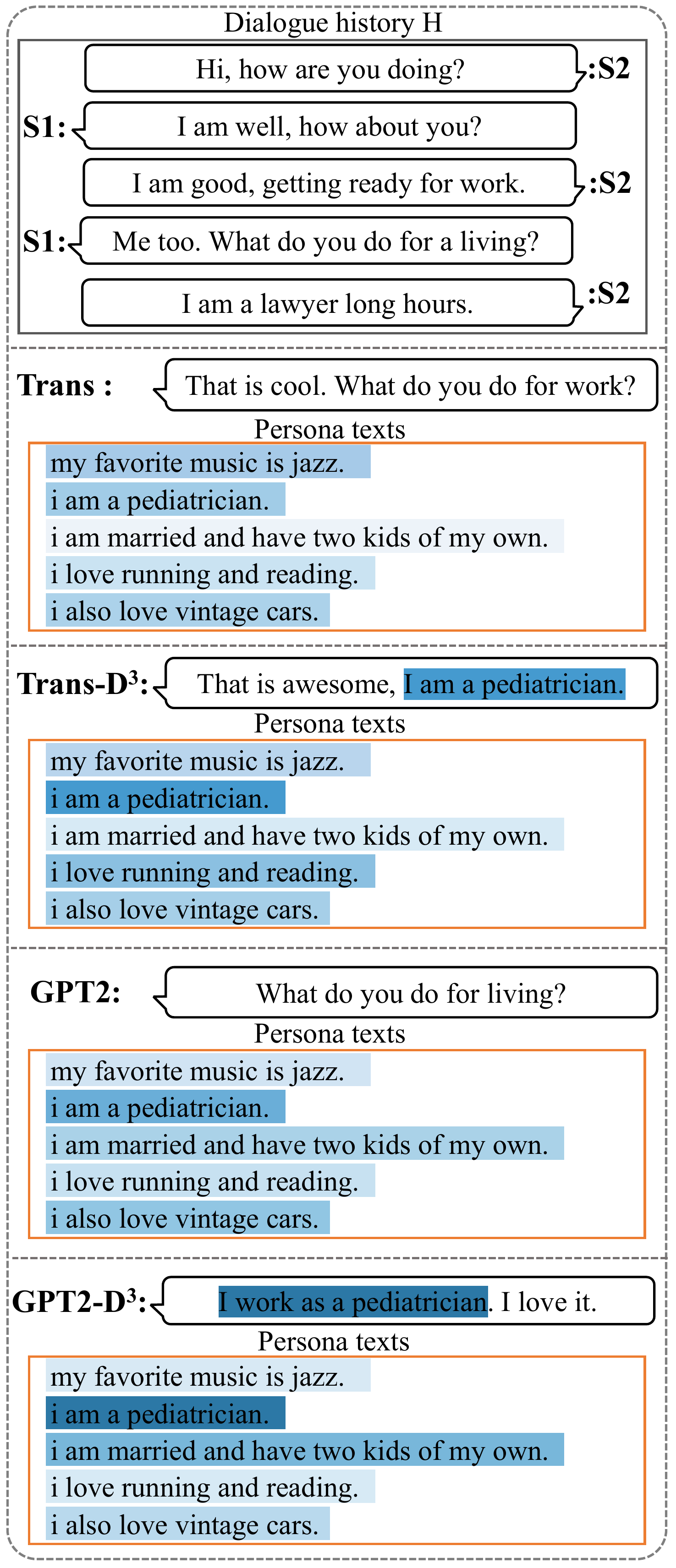}
    \end{subfigure}
    \begin{subfigure}{0.49\linewidth}
        \centering
        \includegraphics[width=1\linewidth]{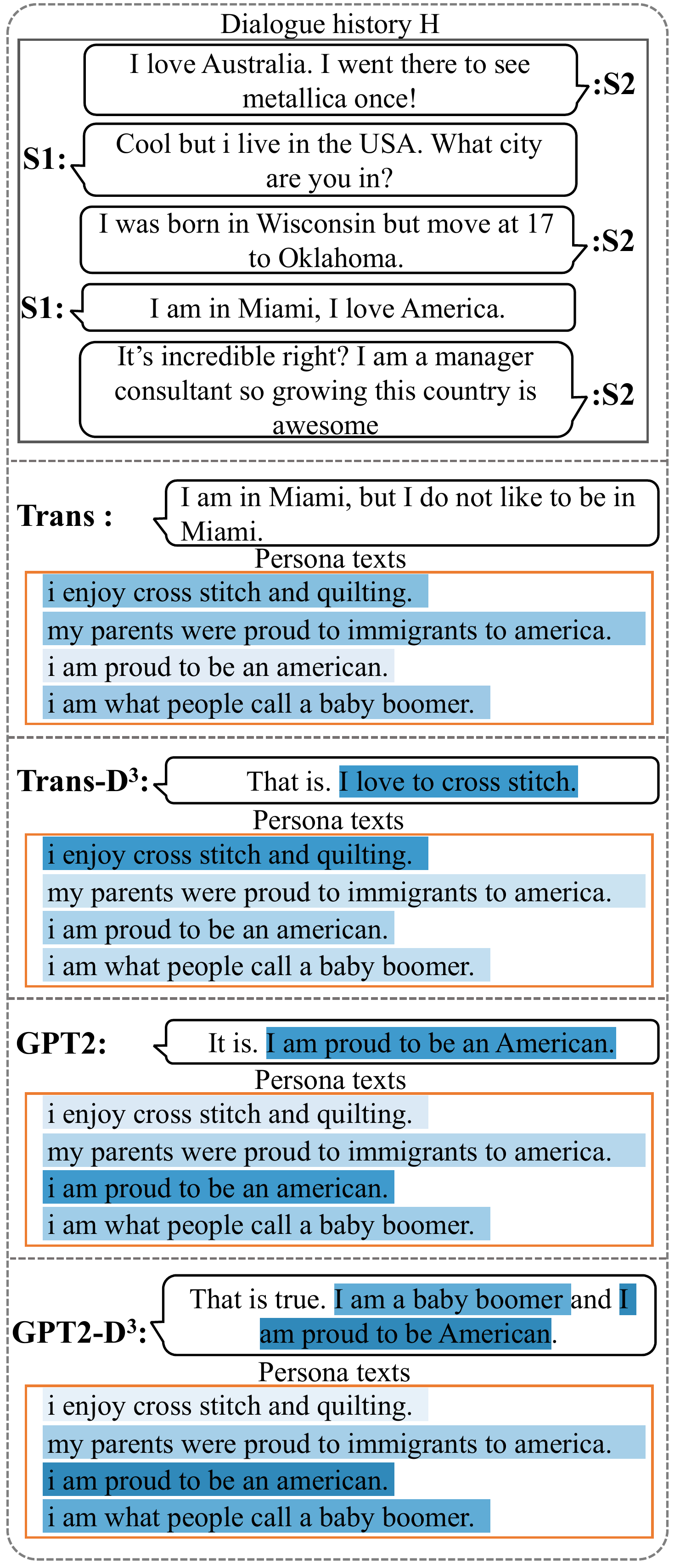}
    \end{subfigure}
    \caption{\label{fig:more_cases} Additional responses cases and visualization by Transformers(Trans) and GPT2 without or with our \textbf{D$^3$} data augmentation method. Colors in each persona text indicate the attention weight paid by different models. A darker color means a higher attention weight is posed by the current model. Colored texts in the response denote the persona consistency.}
\end{figure*}

Except for the cases provided in \secref{sec:more_analysis}, we provide additional cases including the responses given by GPT2. They are shown in Figure~\ref{fig:more_cases}, including visualized attention weights posed by different models on their persona sentences. Note that the attention weights are normalized along the whole input sequence including dialogue history. It can be found that our method can help the model to pay more attention to suitable persona parts, thus the generated responses have better persona consistency.

%The instruction for human annotators during evaluation is shown in Table~\ref{tab:instruction}.

\begin{table*}[t!]
\small
\centering
\setlength{\tabcolsep}{1.4mm}
\begin{tabular}{l|c}
\toprule
\tabincell{l}{Data \\ description} & \tabincell{l}{You are supposed to be \textbf{Speaker S2}, \\ 
you are required to evaluate the \textbf{quality of dialogue responses from S2} in the following 3 aspects, \\ 
based on 1) \textbf{the persona information of S2}; and 2) \textbf{the dialogue history with Speaker S2}. \\
~\\
Here, \textbf{the persona information of S2} mean the personality/characteristics of the speaker for the response \\ 
need to be evaluated. The responses are expected to reflect the given persona for the speaker as possible, \\
meanwhile, they should also be proper and coherent for the previous messages from Speaker S1.\\
~\\
\textbf{Each serial number indicates one sample.} \\
It contains persona information and corresponding dialogue history. \\
The dialogue history contains several different responses (by different methods). \\
Your need to \textbf{rating for every response} considering the persona information and dialogue history.\\ 
~\\
\textbf{Rating contains the following 3 aspects.} } \\
\bottomrule
\end{tabular}

\begin{tabular}{l|l|l}
\toprule
\multicolumn{3}{c}{\textbf{1. Fluency} (1 $\sim$ 3. Your need not to consider the persona information and dialogue history, just the response itself.} \\
\midrule
Score & Description & Examples   \\
\midrule
1 (unsatisfied) & 
\tabincell{l}{1) The text is totally broken, or contains \\ severe grammar errors. \\ 2) The text is very hard to understand} &  
\tabincell{l}{
S1: i do not have any but charlie my puppy enjoys it \\
S2: i am triplets triplets triplets triplets triplets \\
(Cannot understand) \\
~\\
S1: i am a college student . art major .\\
S2: i love my spanish . is studying it has been studying ?  \\
(Totally not fluent)
}\\
\midrule
2 (fair) & 
\tabincell{l}{1) The text is basically fluent, contains \\
grammar errors but do not affect understanding. \\ 
2) The response is short but fluent, without \\
grammar error. \\ 
3) The text contains some repeated context. \\ 
4) The text is bassically fluent, but contains\\
perverse content.} & 
\tabincell{l}{
S1: good , you have any hobbies ?\\
S2: i travel a lot \\
(Fluent but too simple) \\
~\\
S1: what kind of dog is he ? \\
S2: he is a german shepard . he is a german shepard .\\
(Fluent but contains repetitions)
}\\
\midrule
3 (satisfied) &
\tabincell{l}{1) The text is long and informative, few grammar \\ errors are acceptable. There may exist some\\
non-fluent parts, but do not affect understanding. \\ 
2) The text is in medium length, fluent \\ 
without grammar error. } & 
\tabincell{l}{
S1: hello what are doing today ? \\
S2: hello , i just got back from the gym . how are you ? \\
~\\
S1: good , you have any hobbies ? \\
S2: i used to be a painter , i still like to do that \\
a lot . how about you ?\\
} \\
\bottomrule
\end{tabular}

\begin{tabular}{l|l|l}
\toprule
\multicolumn{3}{c}{\tabincell{c}{\textbf{2. Dialogue coherence} (1 $\sim$ 3. You need not to consider the fluency if there is no difficulty in understanding.\\ Your need to consider both the repsonse and dialogue history.)}} \\
\midrule
Score & Description & Examples   \\
\midrule
1 (unsatisfied) & 
\tabincell{l}{The response is irrelevant to the dialogue history. \\ E.g., it does not share the same topic or it is an\\ irrelevant answer. } &  
\tabincell{l}{
S1: how old are you ? i turned four on my birthday !\\
S2: awesome ! i love the insane clown posse love \\
(Irrelevant answer) \\
}\\
\midrule
2 (fair) & 
\tabincell{l}{Very limit relevance exists between the response \\ and history, or meets the following conditions: \\
1) The response is the same as the query. \\
2) The response is a kind of paraphrase of \\ 
the query. \\
3) It is a general response that do not answer the\\
query or contains very limited information,e.g.,\\ "i am sorry"\\
4) The response is a question without new \\ information.} & 
\tabincell{l}{
S1: yes i bet you can get hurt . my wife works\\
and i stay at home\\
S2: i wish i could do that \\
(very limited relevance)\\
~\\
S1: hi ! do you like turtles ?\\
S2: yes i do , do you have any hobbies ? \\
(a question without new information) \\
~\\
S1: i would love to travel to italy . i \\
love baking cookies .\\
S2: i would love to visit italy sometime .\\
(Praphrasing the query)\\
}\\
\midrule
3 (satisfied) &
\tabincell{l}{1) The text is long and informative, few grammar \\ errors are acceptable. There may exist some\\
non-fluent parts, but do not affect understanding. \\ 
2) The text is in medium length, fluent \\ 
without grammar error. } & 
\tabincell{l}{
S1: hello what are doing today ? \\
S2: hello , i just got back from the gym . how are you ? \\
~\\
S1: good , you have any hobbies ? \\
S2: i used to be a painter , i still like to do that \\
a lot . how about you ?\\
} \\
\bottomrule
\end{tabular}
\caption{\label{tab:instruction1}The instruction for annotators to make human evaluation for the generated responses (Part 1).}
\end{table*}

\begin{table*}[t!]
\small
\centering
\setlength{\tabcolsep}{1.4mm}
\begin{tabular}{l|l|l}
\toprule
\multicolumn{3}{c}{\tabincell{c}{\textbf{3. The consistency with given persona} (0 or 1. Your need to consider both the persona sentences and the response.)}} \\
\midrule
Score & Description & Examples  \\
\midrule
0 & \tabincell{l}{The response totally does not\\ reflect any given persona \\ information.} &
\tabincell{l}{Persona sentences: \\
1) i was born in south carolina.\\
2) hey there i am a professional singer.\\
3) i graduated from usc.\\
4) my name is joanna and i love watching horror films. \\
~\\
S2: what is your favorite movie ? (totally irrelevant to persona) \\
~\\
S2: I was born in Texas. So where is your home twon ? \\
("born in Texas" contradict the persona sentence "i was born in south carolina". \\
And there is no other text can reflect the correct persona.)} \\
\midrule
1 & \tabincell{l}{The response can reflect one or \\ several persona sentences directly \\or indirectly.} &
\tabincell{l}{Persona sentences: \\
1) i read twenty books a year.\\
2) i'm a stunt double as my second job.\\
3) i only eat kosher.\\
4) i was raised in a single parent household.\\
~\\
S2: nice . i love to read . \\
(directly reflect the persona "i read twenty books a year.") \\
~\\
S2: nice ! i am currently reading a horror novel . \\
(Indirectly reflect the persona "i read twenty books a year.")} \\
\bottomrule
\end{tabular}
\caption{\label{tab:instruction2}The instruction for annotators to make human evaluation for the generated responses (Part 2).}
\end{table*}

% \end{table*}

% \end{table*}

\end{document}